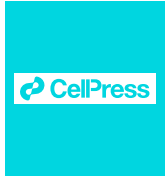
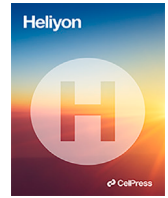

Research article

# Predicting life satisfaction using machine learning and explainable AI

Alif Elham Khan [*], Mohammad Junayed Hasan, Humayra Anjum, Nabeel Mohammed, Sifat Momen

*Department of Electrical and Computer Engineering, North South University, Plot # 15 Block B, Bashundhara R/A, Dhaka, 1229, Bangladesh*

A B S T R A C T

Life satisfaction is a crucial facet of human well-being. Hence, research on life satisfaction is incumbent for understanding how individuals experience their lives and influencing interventions targeted at enhancing mental health and well-being. Life satisfaction has traditionally been measured using analog, complicated, and frequently error-prone methods. These methods raise questions concerning validation and propagation. However, this study demonstrates the potential for machine learning algorithms to predict life satisfaction with a high accuracy of 93.80% and a 73.00% macro F1-score. The dataset comes from a government survey of 19000 people aged 16-64 years in Denmark. Using feature learning techniques, 27 significant questions for assessing contentment were extracted, making the study highly reproducible, simple, and easily interpretable. Furthermore, clinical and biomedical large language models (LLMs) were explored for predicting life satisfaction by converting tabular data into natural language sentences through mapping and adding meaningful counterparts, achieving an accuracy of 93.74% and macro F1-score of 73.21%. It was found that life satisfaction prediction is more closely related to the biomedical domain than the clinical domain. Ablation studies were also conducted to understand the impact of data resampling and feature selection techniques on model performance. Moreover, the correlation between primary determinants with different age brackets was analyzed, and it was found that health condition is the most important determinant across all ages. The best performing Machine Learning model trained in this study is deployed on a public server, ensuring unrestricted usage of the model. We highlight the advantages of machine learning methods for predicting life satisfaction and the significance of XAI for interpreting and validating these predictions. This study demonstrates how machine learning, large language models and XAI can jointly contribute to building trust and understanding in using AI to investigate human behavior, with significant ramifications for academics and professionals working to quantify and comprehend subjective well-being.

## 1. Introduction

Life satisfaction is an essential aspect of human well-being. Research suggests people who are more satisfied with their lives have better mental health outcomes, such as reduced rates of depression and anxiety, are more engaged in their work, and are less likely to experience burnout [1,2]. On the other hand, those with lower levels of life satisfaction are more likely to have poor mental health outcomes and decreased productivity. Researchers in psychology, economics, and other social sciences have focused on life satisfaction for many years. Given the importance of life satisfaction in determining people's overall well-being and quality of life, governments from around the globe are acknowledging its value in policy-making. Governments such as the United Kingdom and Bhutan have created programs and indexes to measure and promote life satisfaction as a vital sign of growth and development [3,4]. The conventional approaches for evaluating life satisfaction predominantly rely on analog approaches, which are inherently time-consuming, expensive, and logistically challenging, especially when targeting large populations. This limitation impedes the widespread adoption of life satisfaction assessments, hindering their utility in effectively informing policy-making and intervention strategies. In contrast, this study leverages machine learning algorithms to predict life satisfaction, thereby transcending the constraints of traditional analog methods. By harnessing the power of machine learning, vast amounts of data can be efficiently processed, enabling the analysis of diverse factors that influence life satisfaction. The utilization of various machine learning models, including decision tree-based and boosting models, underscores the versatility and robustness of our approach in capturing the complexities of human behavior and subjective well-being. Importantly, the methodology of this study achieves high accuracy in predicting life satisfaction, as evidenced by our ensemble models attaining performance scores of 93.80% accuracy and 73.00% macro F1. This level of precision surpasses conventional methods, instilling confidence in the reliability and validity of our approach. Moreover, the 27-item questionnaire used in the study was short and effective, making it highly reproducible, simple, and easy to interpret. It was extracted from a Dutch survey comprising 243 questions using essential feature extraction techniques such as RFECV. The extracted questionnaire only contains the questions that identify the most important factors in detecting contentment in individuals. It sheds light for academics and policymakers on the right questions to ask to assess satisfaction in people's lives. Additionally, ablation studies were conducted to understand the contribution of data resampling techniques and feature selection methods to the overall performance of the machine learning models, highlighting the importance of the dual balancing strategy employed in this study, as well as the effectiveness of the RFECV-based feature selection approach in extracting the most salient determinants of life satisfaction.

[*] Corresponding author.
*E-mail address:* alif.khan1@northsouth.edu (A.E. Khan).

In addition to traditional machine learning techniques, the research delved deeper into the use of large language models (LLMs) such as BERT, BioBERT, ClinicalBERT, and COReBERT for the purpose of predicting life satisfaction. This was accomplished after a thorough LLM-specific data preprocessing to generate natural language texts from tabular data. The results of these experiments provide valuable insights into the suitability of different LLMs for this domain, with BioBERT demonstrating the strongest performance and suggesting that life satisfaction prediction is more closely aligned with the biomedical domain than the clinical domain.

The study also compared the primary determinants across different age brackets to explore how the appropriate questions and determinants change throughout life. Moreover, the adoption of explainable AI (XAI) ensures transparency and interpretability, elucidating the rationale behind each prediction and empowering stakeholders to make informed decisions based on actionable insights. XAI enables the creation of machine learning models that are more transparent, interpretable, and understood by humans, potentially increasing their reliability and trustworthiness. This aspect is especially crucial for decision-makers, such as policymakers, who must comprehend and apply the results of such models to establish effective initiatives that enhance well-being and mental health. By providing transparency, interpretability, actionable insights, risk assessment, and customization capabilities, explainable AI demystifies the decision-making process and enables decision-makers to understand, validate, and act upon the model's predictions effectively. Through its ability to translate complex machine learning models into understandable terms, explainable AI empowers decision-makers to make informed judgments, drive positive outcomes, and navigate the complexities of decision-making in diverse domains. The major accomplishments of this study can be summarized as follows:

- The study identified the primary determinants influencing individuals' life satisfaction.
- The dominant determinants were used to develop a simple and efficient questionnaire for assessing people's levels of contentment.
- Several machine learning algorithms have been applied to create models that achieve a high level of accuracy in predicting the state of contentment in individuals.
- Textual data was used to predict life satisfaction, with large language models (LLMs). Textual data was generated by transforming tabular data using the techniques of mapping and concatenating categorical data to create meaningful text data. LLMs complement the traditional machine learning approaches.
- Explainable AI techniques were employed to enhance the interpretability of the outcomes in our AI-based models, rendering them more accessible to human understanding.
- The primary determinants were compared and analyzed across four age brackets.
- Deployment of our interactive app enables individuals to predict their state of contentment, expanding the accessibility of our research and providing a practical tool for well-being assessment.

## 2. Related works

Life satisfaction surveys were first conducted in the US in the 1960s, with subsequent studies in different countries evaluating various factors that affect overall life satisfaction, such as mental health [5,6], age [7], socioeconomic factors [8] and many more [9–14]. Several nations are collecting national life satisfaction statistics [15], and research is being conducted on the impact of environmental factors [16], social media [17], disability discrimination [18], and personality [19] on life satisfaction. Longitudinal panel studies have been conducted in Australia and Germany to track life satisfaction over time.

Various country-wise studies and evaluations of life satisfaction were carried out over the past few decades. The most elementary use of life-satisfaction data is to estimate the apparent quality of life within a country or a specific social group [20]. The USA, for example, has focused a lot and given significant importance to this field. Much emphasis has been shown on mental health in many previous types of research [5,6]. In the 1970s, life satisfaction was a central theme in American society. Indicator studies. Landmark books were published stating satisfaction with life, as a whole, is calculated based on satisfaction with various aspects of life [21,22]. Some recent works in the USA presented a few updated methodologies in determining life satisfaction. The Center for Disease Control and Prevention in the USA measures life satisfaction in particular of its large health surveys [23].

The United Kingdom, on the other hand, is collecting national life satisfaction statistics for possible policy use [15]. Life satisfaction was studied extensively from the perspectives of social media [24], sexual minority [15], disability discrimination [18], personality effects [19] and environmental factors [25]. Life satisfaction from a social media perspective was carried out using random-intercept cross-lagged panel models. The sexual minority study used data to estimate a simultaneous equations model of life satisfaction. The model allows for self-reported sexual identity to influence a measure of life satisfaction directly and indirectly. The disability discrimination study focused on prospective associations between disability discrimination and well-being through an extensive cohort study. The personality effect study was carried out by testing cross-sample affect replication.

Japan [26] and Chile [27] are also taking measures to evaluate life satisfaction. While Japan evaluates life satisfaction using prospective cohort studies, Chile measures satisfaction for seven life domains by combining deductive and inductive approaches.

Other nations, such as Germany and Australia, have ongoing longitudinal panel studies in which life satisfaction is tracked over time. Germany intended to evaluate life satisfaction based on economic [28], environmental [16] and educational [29] perspectives through decomposition analysis, empirical analysis of probit model, and bottom-up analysis, respectively.

Regarding a sense of subjective well-being, researchers have asserted that it consists of two main components: the emotional component and the cognitive component [30,31]. The cognitive part has been more closely conceptualized with life satisfaction [21]. Despite this, it had not previously received much attention for research. Diener and teammates sought to address this, and through developing the SWLS, they created a vital tool for measuring the cognitive components they felt reflected a subjective sense of well-being and life satisfaction.

Until now, none of the mentioned works has used any machine learning algorithm to try to predict life satisfaction. Very recent work in 2022 in this field included using machine learning to uncover the relation between age and life satisfaction [32]. Another research in 2022 focused on Machine learning techniques to identify the correlates of quality of life [33]. Lastly, another work in 2015 has taken advantage of social media data to evaluate SWL (satisfaction with life and SWB (subjective well-being) where data-driven supervised learning methods were used to predict SWL from a set of features extracted from Facebook [17]. Table 1 provides a detailed analysis of previous works in this field, including their contributions and limitations.

## 3. Materials and methods

### 3.1. Objectives

This study intends to approach and unravel a multifaceted view of human contentment, guided by a set of research questions. These questions were designed to interweave the realms of human psychology with the precision of machine learning in pursuit of a deeper understanding of life satisfaction and ways to measure it accurately. The methodology that was employed aligns neatly with our objectives, each of which addresses a corresponding research question.

1. **Identification of Life Satisfaction Determinants:** In response to the first research question, *"What are the most significant determinants influencing life satisfaction among individuals, and how do these factors interrelate?"*, the primary objective of this study was to identify these determinants systematically. Recursive Feature Elimination with Cross-Validation (RFECV) was deployed on our comprehensive dataset, narrowing the selection from an initial pool of 243 variables to the most critical 27. The method allows us to identify these pivotal factors and gain insight into their interrelationships and relative importance.
2. **Optimization of Contentment Prediction:** Guided by the second research question, *"How effectively can an ensemble of machine learning algorithms predict an individual's state of contentment, and what are the primary indicators contributing to this accuracy?"*, the second objective of this study was the development of an accurate prediction model. An ensemble of machine learning algorithms was employed to combat the imbalanced nature of our dataset and maximize both the accuracy and the F1 score. This choice attests to our commitment to ensuring the validity and robustness of the contentment prediction model of this study.
3. **Amplification of Model Interpretability with Explainable AI:** The last objective of this study, stemming from the third research question *"In what ways can Explainable AI methodologies enhance the interpretability of AI-based models utilized for predicting personal life satisfaction, and what impact does this enhanced understanding has on the application and acceptance of these models?"* sought to bring a layer of transparency to the AI model. This was done through the employment of Explainable AI, a technique designed to unveil the underlying decision processes of the model. Two case studies were presented—one in which the model predicts contentment and another in which it does not exhibit the Explainable AI's capacity to make our models understandable and trustworthy.

In pursuing these objectives, a multi-pronged approach was employed that combines advanced machine learning methods with psychological insights, aiming to bring a novel, practical contribution to understanding and measuring life satisfaction.

**Table 1**
A comprehensive analysis of the existing literature in life satisfaction prediction. We divide the works into three categories and provide in-depth analyses of the main contributions and limitations of the works aimed to be handled in our research.

| Methods of Data Collection | Ref. | Subjects | Area | Main Contribution | Limitation |
|---|---|---|---|---|---|
| National Survey | [21, 22] | 5000+ adults | Different states of the USA | Provide a comprehensive framework for measuring well-being using objective and subjective indicators | Self-reported data, might be biased or inaccurate |
| Personal Interview | [7] | 1500 adults | USA | Establish correlates of life satisfaction among older persons | Small sample size, not generalizable |
| Interview survey | [34] | 10000+ individuals | Scandinavian countries | Use 'welfare state' to represent life satisfaction. | Complex method of surveying |
| *Statistical Analyses* | | | | | |
| Conceptual Analysis | [35] | N/A | N/A | Argue for a broader understanding of happiness that includes life satisfaction as a crucial component | No empirical analysis |
| Comprehensive Review | [23] | People from all ages, sex, and gender | More than 150 nations | Synthesis of existing knowledge on life satisfaction scales | Does not explore the relationship between objective and subjective measures of well-being |
| Regression, Mediation Analysis | [15] | 45352 people of all ages | Australia and UK | Provide empirical evidence on the relationship between sexual orientation and life satisfaction | Biased and unable to establish causality |
| Fixed-effects modeling, Sensitivity analysis | [24] | 12672 adolescents aged 10 to 15 years | UK | Established that relationship between social media use and life satisfaction is small and often non-existent | Biased, does not consider potentially harmful effects of social media |
| Prospective cohort study | [18] | 871 people with self-reported physical, cognitive or sensory disability | UK | Established that experiencing disability discrimination is associated with a decline in well-being among older adults | Complex methodology, limited research domain and biased |
| Regression Analysis | [25] | 40000+ survey participants | European countries | development of a new approach to environmental valuation that relies on individuals' life satisfaction | Relies on assumption and unsuitable approaches |
| Cross-sectional analysis | [16] | 20000+ individuals | Germany | Provides evidence that the socioeconomic status of a neighborhood has a significant impact on individuals' life satisfaction | Causality cannot be inferred due to the cross-sectional design of the study. |
| Hedonic approach | [29] | 33395 individuals of various ages and genders | Germany | Finds a positive and significant relationship between local environmental quality and life satisfaction | Relies on self-reported data and does not account for the potential endogeneity of environmental quality. |
| Conceptual framework model | [36] | Survey participants from 1981 to 1987 | Australia | Theorize that subjective life satisfaction is fairly stable in financial context | Limited empirical evidence and lacks specific measures. |
| *Machine Learning Approaches* | | | | | |
| Machine learning algorithms | [37] | 2853 individuals | Turkey | Detect anxiety state using socio-economic data with model interpretability | May contain linguistic or contextual inaccuracies due to the translation of the original Turkish dataset. |
| Machine learning and deep learning | [38] | 684 students aged 19 to 35 | Bangladesh | Developed a dataset, a hybrid depression assessment scale and provided model interpretability for depression assessment | Applicability is constrained to the demographic and cultural context, potentially affecting the generalizability of the predictive models across different populations and cultures. |
| Machine Learning, Neural Networks | [32] | 400,000 observations from SOEP survey | Germany | Assess the relation between life satisfaction and age | Based on cross-sectional data. Limits the ability to make causal inferences about the relationship between age and life satisfaction. Unexplained model prediction |
| Multiple machine learning algorithms | [33] | 400,000 observations from SHARE survey | European countries | Identify the correlates of quality of life | Limited to European adults older than 50. Unexplained model prediction. |
| Machine Learning Algorithms | [17] | 58000 Facebook users | All over the world | Shows that social media data can be used to predict individual levels of life satisfaction. | Limited to Facebook users, which may affect the generalizability of the findings, sparsely recorded and unexplained predictions. |

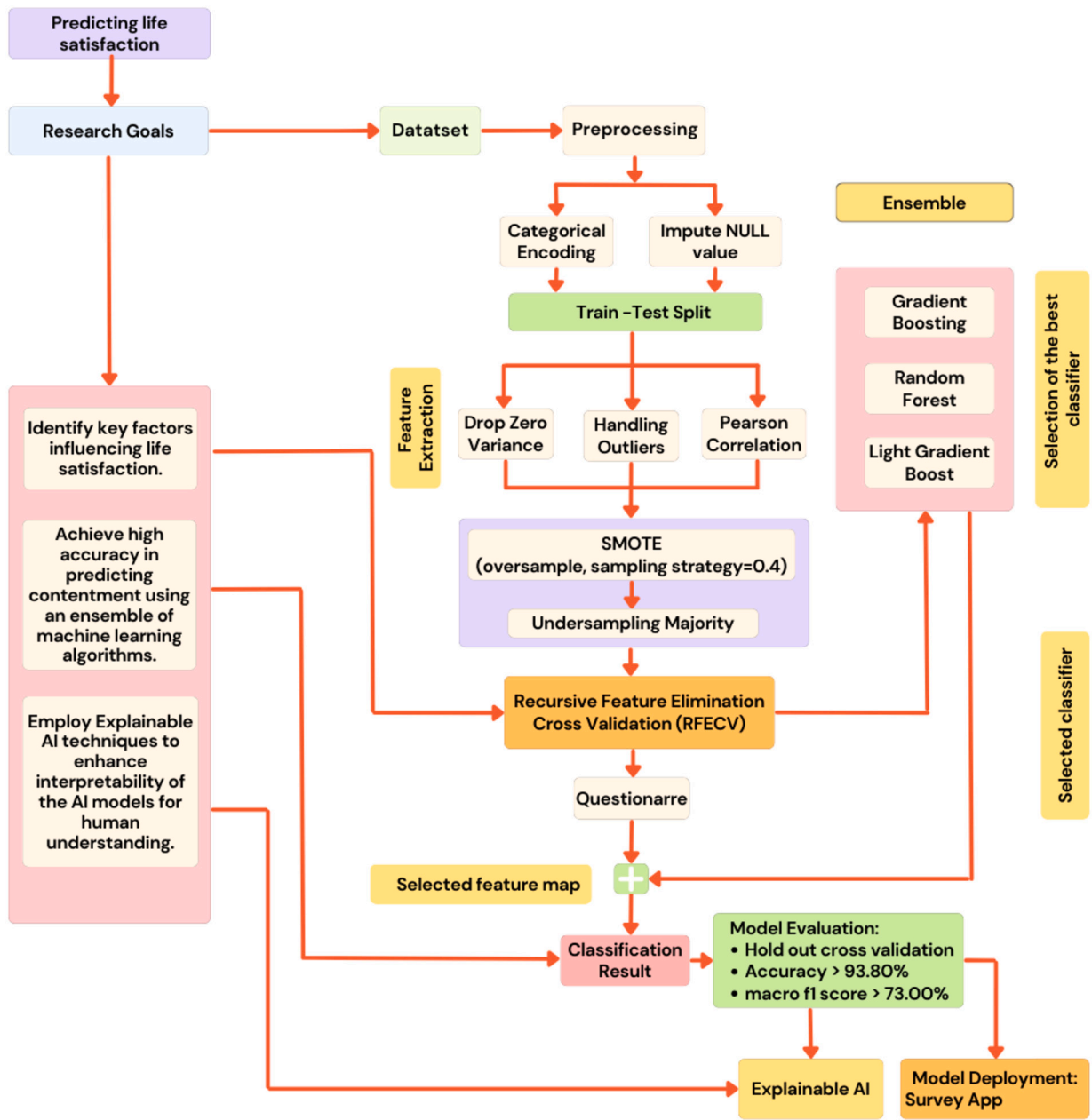

**Fig. 1.** Workflow diagram that outlines the key steps used in applying machine learning algorithms, addressing research objectives, and highlighting the main evaluation results.

### 3.2. Research workflow

Machine learning approaches for clinical psychology and psychiatry explicitly focus on learning statistical functions from multi-dimensional data sets to make generalizable predictions about individuals [39]. In this study, a multidimensional dataset and several state-of-the-art machine learning algorithms were used to learn the patterns in detecting well-being and life satisfaction among individuals. The proposed methodology is shown in the workflow diagram in Fig. 1.

### 3.3. Dataset collection

The dataset serving as the foundation for this research was diligently sourced from a comprehensive survey carried out by the Government of Denmark. This survey, named SHILD (Survey of Health, Impairment and Living Conditions in Denmark), published in 2018 [40], provides robust and diverse information amassed from an extensive pool of participants. The Danish National Centre collects the SHILD data for Social Research (SFI) and is subject to the Danish Data Protection Act and the EU General Data Protection

**Table 2**
Demographic profile of respondents.

| Attributes | Category | Frequency | Percentage (%) |
|---|---|---|---|
| Age | 16-20 | 1654 | 9.8 |
|  | 21-25 | 1163 | 6.9 |
|  | 26-30 | 973 | 5.7 |
|  | 31-35 | 1243 | 7.3 |
|  | 36-40 | 1661 | 9.8 |
|  | 41-45 | 1464 | 8.6 |
|  | 46-50 | 2177 | 12.8 |
|  | 51-55 | 2099 | 12.5 |
|  | 56-60 | 2208 | 13.0 |
|  | **61-64** | **2329** | **13.7** |
| Gender | Male | 7989 | 47.1 |
|  | **Female** | **8982** | **52.9** |
| Employment Status | **Employed** | **11793** | **69.5** |
|  | Unemployed | 4728 | 30.5 |
| Long-term Health Condition | Yes | 3941 | 23.2 |
|  | **No** | **13028** | **76.8** |
| Marital Status | **Married / Partnered** | **13373** | **78.8** |
|  | Single | 3597 | 21.2 |

Regulation. However, the data was publicly released in the study in 2018, making it usable for researchers worldwide. The study focuses specifically on disability, disability, and physical and mental health but also contains information on the panelists' education, employment, loneliness, family situation, violence, discrimination, community participation, and experiences of municipal case management.

The dataset was collected in 2012 and was based on web questionnaires with telephone follow-up questions. The respondents were all Danish citizens, and the response rate of the survey was approximately 50%. The demographic information of the dataset is presented in Table 2.

The survey encompasses responses from approximately 19,000 individuals, a substantial sample size that enhances the reliability and generalizability of our findings. The age bracket of the respondents ranged from 16 to 64 years, offering a broad perspective across different life stages and experiences. This dataset captures a snapshot of critical factors affecting life satisfaction in general. Utilizing this dataset in this study strengthens our commitment to basing our research on comprehensive, representative, and high-quality data. The Danish government's meticulous collection process aligns with our endeavor to deliver reliable, significant, impactful life satisfaction and contentment findings.

*3.4. Data pre-processing*

Quality data is pivotal to model performance in machine learning, requiring a rigorous pre-processing pipeline to ensure predictive reliability. This research utilized a methodical approach to manage and optimize the collected dataset using the following rigorous data pre-processing steps:

*3.4.1. Handling missing values*

The original dataset contains a lot of missing values. Addressing these missing values is imperative, as many machine learning algorithms cannot accommodate them. All the features in the raw dataset comprising more than 20% of null values were dropped. Fig. 2a shows a matrix of missing values where each column represents a feature and each white spot represents a missing value in the corresponding column. Fig. 2b shows the matrix of missing values after dropping the attributes composed of more than 20% null values. The remaining null values were iteratively imputed. In iterative imputation, missing values are sequentially predicted for each feature, beginning with the one having the fewest missing values. Employing Bayesian Ridge regression, these predictions are made based on the available values of other features. This iterative process continues until all missing values are filled. Furthermore, imputing missing values in columns with a moderate percentage of null values (less than 20%) ensures that the majority of the dataset is retained, preserving the overall integrity of the data. Iterative imputation, in particular, takes into account the relationships between variables, potentially leading to more accurate imputed values. By imputing missing values iteratively, this study aims to provide the most accurate estimates possible, which can lead to improved model performance.

*3.4.2. Categorical encoding*

Categorical encoding converts categorical data into integer format so that the machine learning models relying on mathematical formulas can handle the data. Most of the categorical features of our dataset were ordinal. An ordinal encoder was used to maintain the ordinality of the data. Categorical responses were converted into a standardized numerical format. For example, health conditions were encoded using a predefined scale ('*Very well*' = 3, '*Well*' = 2, down to '*Very poor*' = 0).

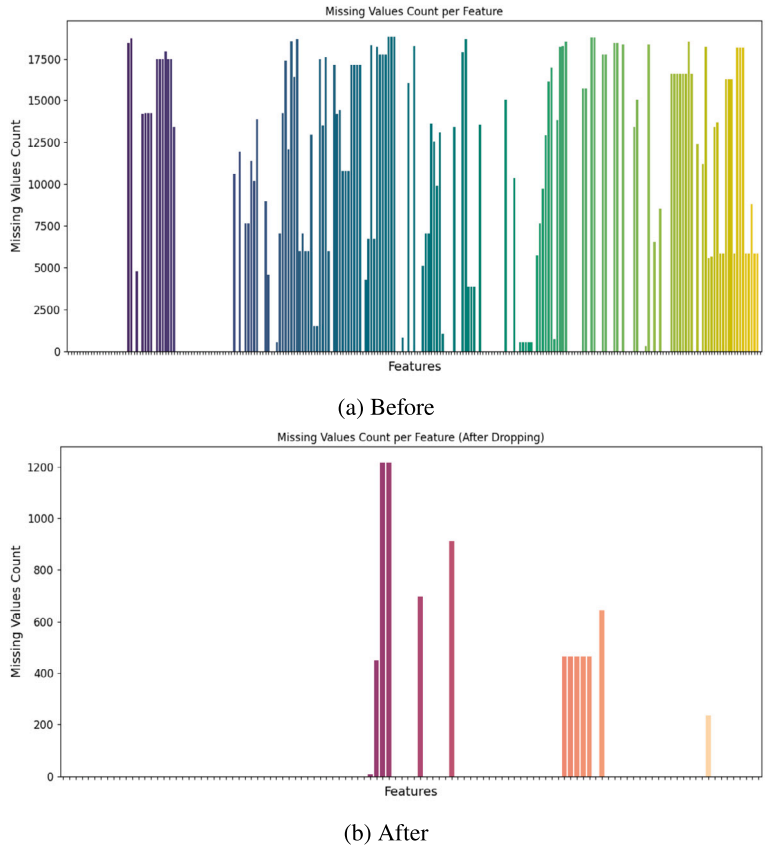

(a) Before

(b) After

**Fig. 2.** Handling missing values. (a) shows several features having a substantial number of missing values before being handled. (b) Only a few features containing null values remain after dropping some features and imputing null values.

*3.4.3. Train test split*

The dataset is split in an 80-20 ratio where 80% of the data is used for training, and the remaining 20% is preserved only for validating the predictions made by our learned model. The dataset was shuffled before splitting it, which is generally a good practice to avoid any potential biases in the data ordering.

*3.4.4. Zero variance*

Features with zero variance were removed because they do not provide useful information for predictive modeling or data analysis.

*3.4.5. Handling outliers*

For each feature, values lying twice their standard deviation from their mean are considered outliers. The medians of respective features replaced the outliers. Equations (1) and (2) show outliers Z are values that are two times the standard deviation away from the mean. These values are replaced with the median. The outlier handling process is shown in Fig. 3a and Fig. 3b.

$$Z < \overline{X} - 2\sigma \qquad (1)$$

$$Z > \overline{X} + 2\sigma \qquad (2)$$

*3.4.6. Resampling imbalanced data*

The target outputs are imbalanced in the dataset, so the study uses a resampling strategy to balance the two strata: "Content" and "Discontent." While doing so, it was ensured that samples from each stratum are represented. Empirically, we find that in our machine learning pipeline, balancing the outputs results in the machine learning model to predict with higher accuracy. The empirical results are reported in Sec. 4.5.1. The study uses a dual balancing strategy where first SMOTE (Synthetic Minority Over-sampling Technique) [41] is used to resample the minority class to 40% of the majority class. Next, the number of samples of the majority class is reduced to that of the minority class using undersampling. The class distribution before and after resampling is shown in Fig. 4a and Fig. 4b.

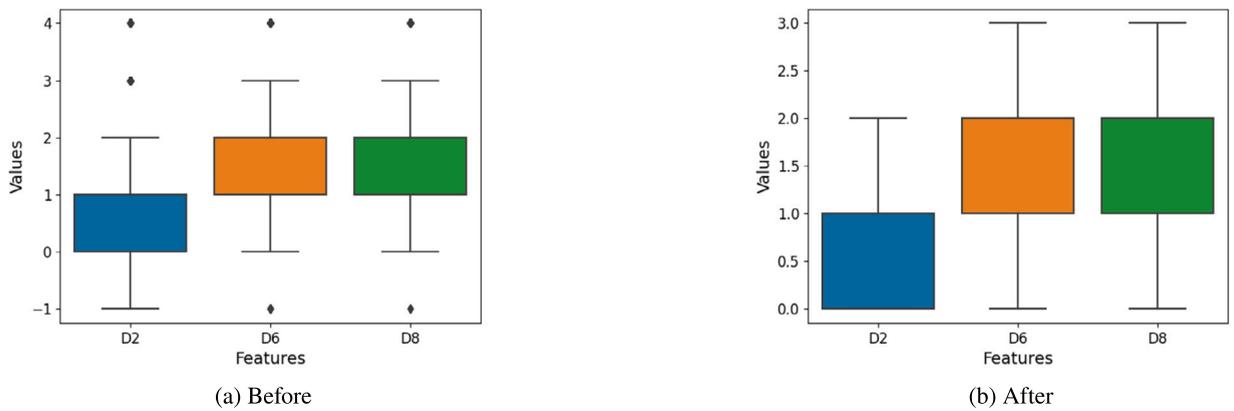

(a) Before        (b) After

**Fig. 3.** Box plots visually showing the handling of outliers. The box plots in (a) show the data distribution before handling the outliers. The box plots in (b) show that the outliers have been handled.

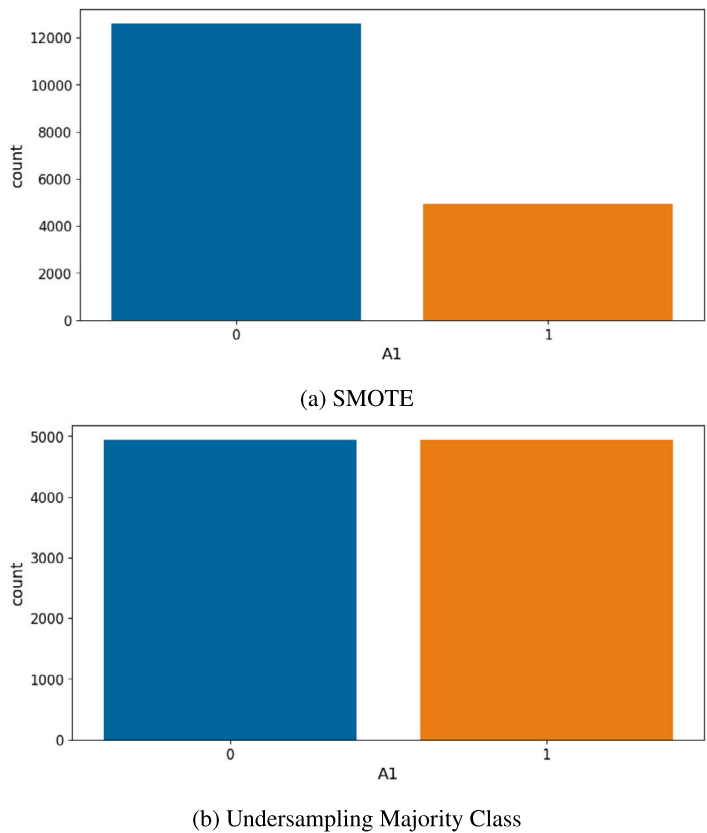

(a) SMOTE

(b) Undersampling Majority Class

**Fig. 4.** Bar plots visualizing the class distribution of the classification task after applying the resampling techniques. (a) shows the distribution after applying the oversampling strategy using SMOTE. (b) shows the distribution after applying the undersampling strategy.

### 3.4.7. Feature selection

The study uses a wrapper feature selection method called RFECV (Recursive Feature Elimination and Cross-Validation Selection). It is an algorithm that eliminates irrelevant features based on validation scores. Random Forest algorithm was used as the estimator for this recursive algorithm. A 5-fold cross-validation was employed. As a result, 27 of the most essential features were extracted. These features have been used to produce the final questionnaire with 27 questions. The following RFECV curve in Fig. 5 represents the performance of a model as a function of the number of selected features during the Recursive Feature Elimination with Cross-Validation (RFECV).

The RFECV feature importance in Fig. 6 provides valuable insights into the importance of each feature in contributing to the overall predictive performance of the model. The height or color of each bar or cell represents the feature's relative importance, allowing us to identify the most influential features in the dataset easily. Here, feature A2 (How would you rate your health generally?) has the highest importance among all other features.

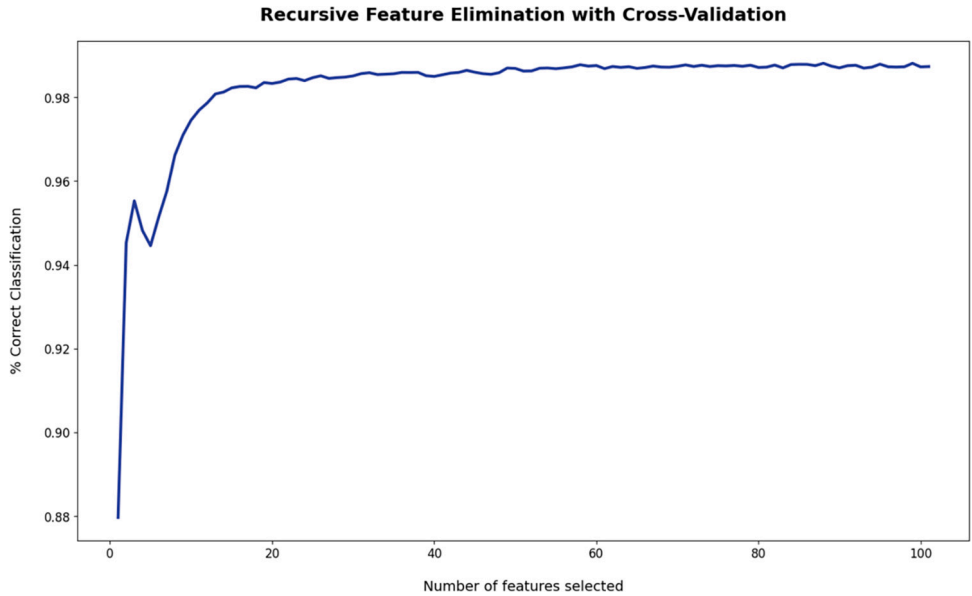

**Fig. 5.** RFECV curve shows the number of important features vs accuracy. More than 98% accuracy is achievable when there are more than 20 important features.

*3.4.8. LifeWell survey*

Fig. 7 shows the 27 extracted research questions after data pre-processing and feature selection steps. The questions have been further categorized according to different aspects of life. Here, the labels (e.g., A2, C1) represent the code of each question as it appears in the original dataset. This compact questionnaire gives us a perspective of the most important questions for analyzing people's contentment.

*3.4.9. Data preparation and pre-processing for LLMs*

Data preparation and preprocessing are critical steps in the pipeline of utilizing large language models (LLMs) for text-based analysis. The following points outline the steps applied to the LifeWell survey dataset to prepare it for analysis using LLMs:

- **Mapping:** In Sec. 3.4.2, the categorical data in our tabular dataset was transformed to a numerical format using an ordinal encoding approach. For creating sentences from encoded tabular data, the encoded values were converted back to textual categorical data, with additional meaningful texts around them, allowing us to generate one meaningful sentence per data instance. This was accomplished using a mapping function that is applied to each column of the dataset, generating 27 meaningful chunks for each of the 27 factors in the LifeWell survey. Two examples of such mapping are shown in Fig. 8.
- **Sentence Generation:** After the mapping, each of the 27 chunks was concatenated to create one single sentence per instance. The resulting sentences were stored in a new DataFrame column, serving as the input for LLMs along with their corresponding labels. Two examples of such generated sentences can be seen in Fig. 8.

The transformation of structured data into natural language sentences aligns with the input requirements of LLMs, such as BERT and its derivatives. This approach optimizes the data format for natural language processing, enhancing both the performance and interpretability of the models.

*3.5. Machine learning algorithms*

*3.5.1. Random forest*

Random Forest (RF) is an ensemble classifier that solves regression and classification problems. It employs several decision trees as base classifiers on various subsets of the given dataset and takes the average to improve the predictive accuracy of that dataset by majority vote. In this experiment, the machine learning model created with this machine learning algorithm produced the confusion matrix in Fig. 9.

*3.5.2. Gradient boosting*

Gradient Boosting is a greedy algorithm that minimizes overall prediction error by relying on the intuition of the best possible next model combined with previous models (eq (3)). It is suitable for minimizing the bias error of a model. The confusion matrix was obtained in Fig. 10 using Gradient Boost.

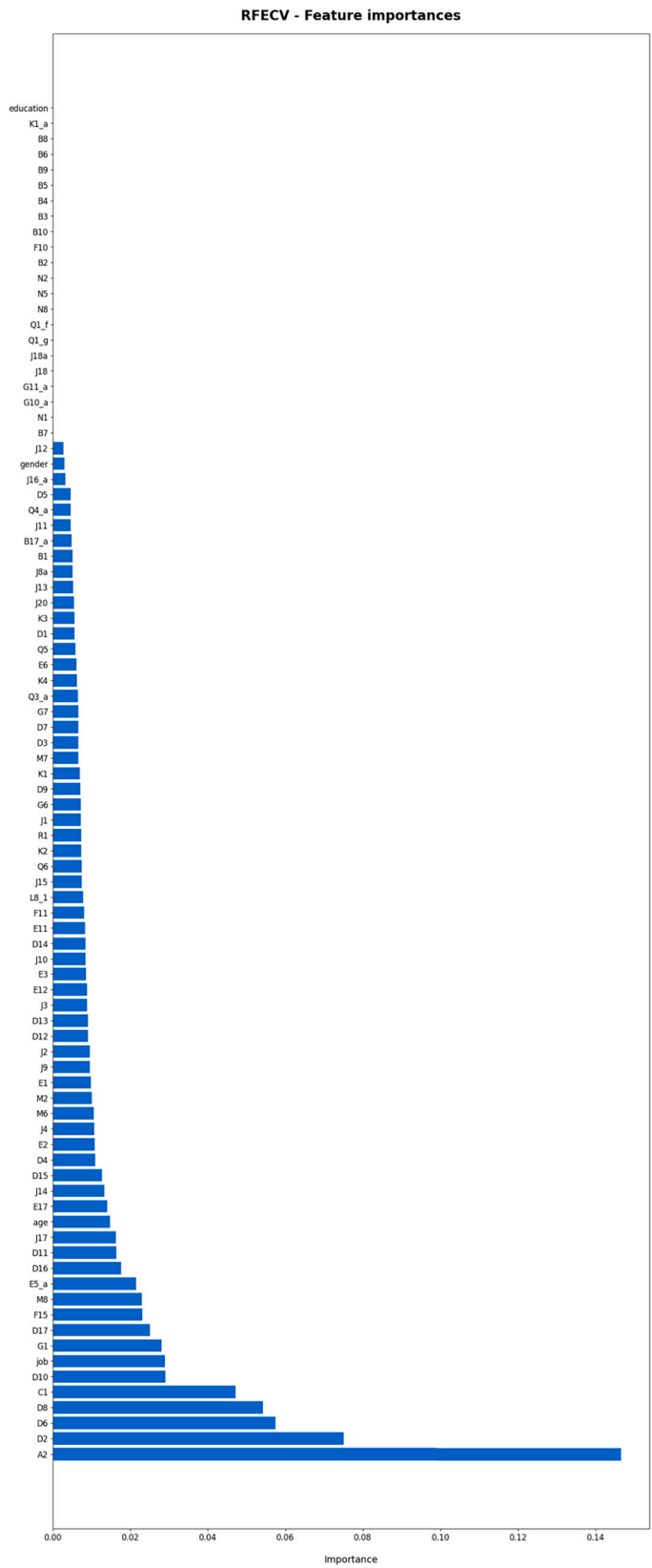

**Fig. 6.** Feature importance obtained from RFECV.

| PHYSICAL | | MENTAL |
|---|---|---|
| What is your age? (age) | | Are you depressed? (D2) |
| How would you rate your health generally? (A2) | | Do you see yourself as a person who is relaxed and handles stress well? (D4) |
| Do you suffer from a long-term physical health problem or disability? (C1) | | Do you see yourself as a person who can be tensed? (D6) |
| | | Do you see yourself as a person who worries a lot? (D8) |
| How tall are you? (number of centimetres) (E1) | | Do you see yourself as a person who does not give up until the task is completed? (D11) |
| How much do you weigh? (kilos) (E2) | | Do you see yourself as a person who is emotionally stable, and not easily excited? (D10) |
| In the past year, have you consulted other practitioners or therapists? (E5 a) | | Do you see yourself as a person who prepares plans and implements them? (D15) |
| | | Do you see yourself as a person who gets nervous easily? (D16) |
| | | Do you see yourself as a person who is easily distracted? (D17) |
| (a) | | (b) |

| ECONOMIC | | SOCIAL |
|---|---|---|
| Do you hold a job? (job) | | Who do you primarily talk to about personal and serious problems? (E17) |
| On a scale from 0-10, where 0 signifies very low and 10 signifies very high, how content are you generally with your job? (F15) | | Do you have a spouse/partner/boy/girlfriend? (G1) |
| What is your primary source of income, when considering all your sources of income? (M2) | | How often have you spent time with other relatives in the past year? (J2) |
| In the past year, how much have you spent on medicine, nutritional supplements? (M6) | | In the past year, how often have you spent time with acquaintances? (J4) |
| How would you rate your current finances? (M8) | | How often have you been abroad on holiday or family visits in the past year? (J17) |
| (c) | | (d) |

| CULTURAL |
|---|
| In the past year, how often have you been to the cinema, a concert, or the theater? (J9) |
| How often have you read a newspaper in the past year (not counting online newspapers)? (J14) |
| (e) |

**Fig. 7.** The survey questionnaire on multidimensional aspects of life as produced by our study. These questions from different aspects of life represent the key indicators of life satisfaction, a crucial finding of this research. (a) shows the indicators related to physical health. (b) shows the indicators related to mental and psychological health. (c) shows the economic indicators. (d) shows the social indicators, and (e) shows the cultural indicators playing the most significant role in life satisfaction.

$$\hat{y}(x) = \sum_{t=1}^{T} \eta_t h_t(x) \tag{3}$$

### 3.5.3. Decision tree

A Decision Tree is a supervised machine-learning algorithm used to solve classification problems. In this experiment, the machine learning model created with this machine learning algorithm produced the confusion matrix in Fig. 11.

### 3.5.4. AdaBoost

AdaBoost or Adaptive Boost is an ensemble learning method that can be used for both classification and regression problems. This technique combines weak classifiers into a robust classifier (eq (4)). The confusion matrix in Fig. 12 was obtained using AdaBoost in our pipeline.

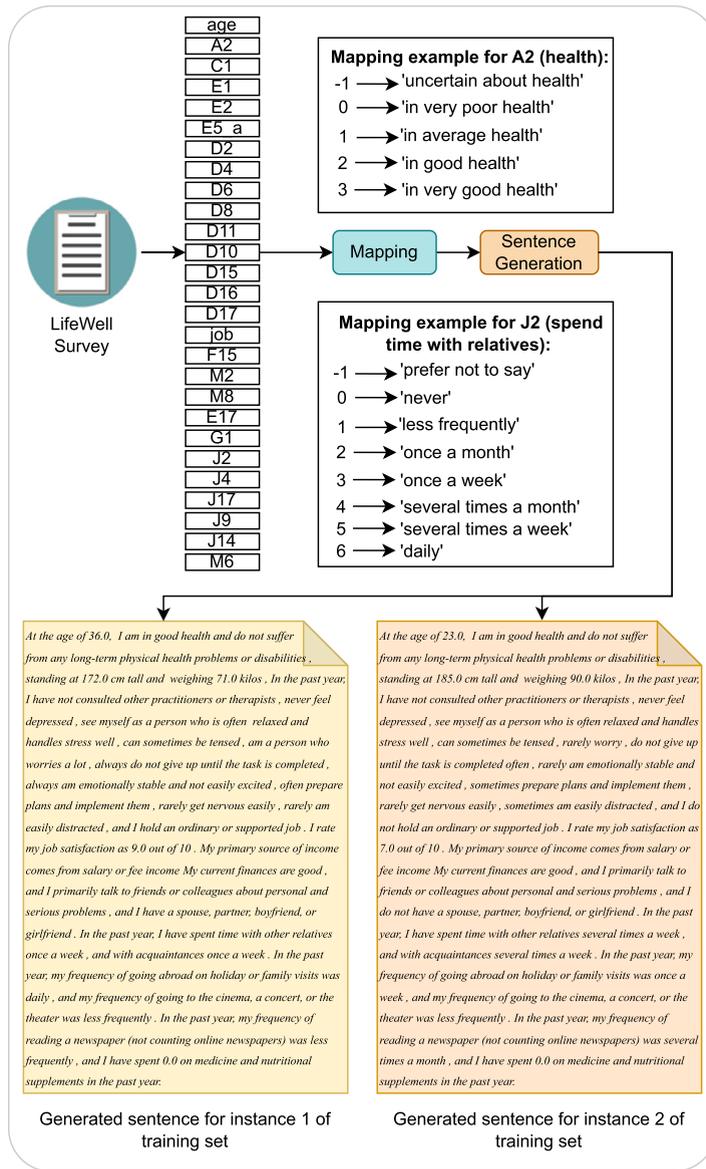

**Fig. 8.** The pre-processing steps performed in order to generate sentences from tabular data of the LifeWell survey generated by our feature selection methods. The categorically encoded numeric values were converted back to the original representation with additional meaningful words to generate the sentences to be fed into LLMs for classification.

$$\hat{y}(x) = \text{sign}\left(\sum_{t=1}^{T} \alpha_t h_t(x)\right) \qquad (4)$$

### 3.5.5. XGBoost

XGBoost is an implementation of gradient boosting used for supervised learning problems (regression, classification, ranking, etc.). XGBoost creates decision trees in sequential form, and the individual classifiers $f_{k(x)}$ give a precise model when they are an ensemble (eq (5)).

$$\hat{y} = \sum_{k=1}^{K} f_k(x) \qquad (5)$$

The confusion matrix in Fig. 13 was obtained from XGBoost.

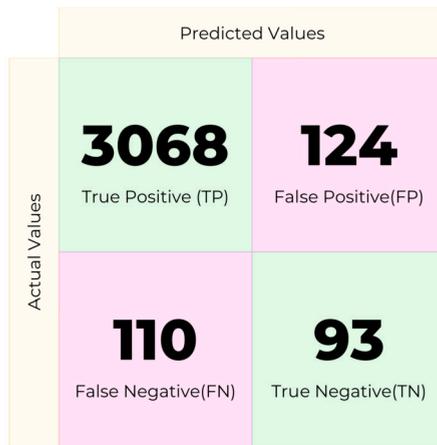

**Fig. 9.** Confusion matrix for Random Forest.

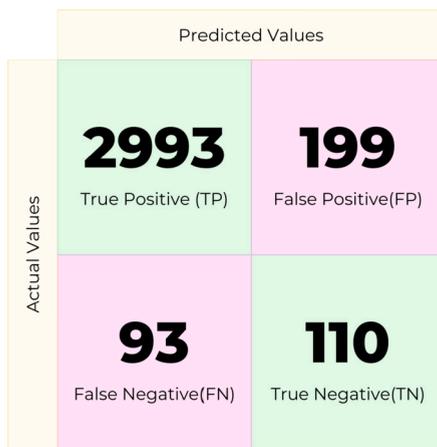

**Fig. 10.** Confusion matrix for Gradient Boosting.

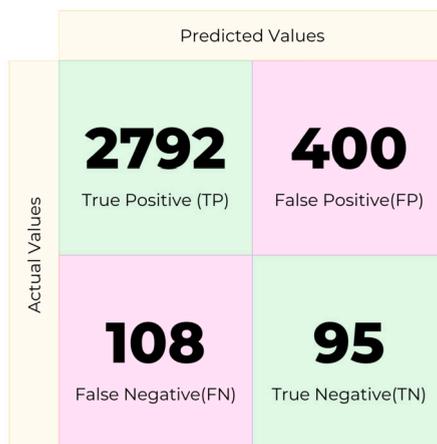

**Fig. 11.** Confusion matrix for Decision Tree.

#### 3.5.6. SVC

SVC (Support Vector Classifier) is a binary classification algorithm that identifies the optimal hyperplane for splitting classes. It optimizes the distance between the hyperplane and the nearest data points, referred to as support vectors (eq (6)). The confusion matrix obtained from SVC is shown in Fig. 14.

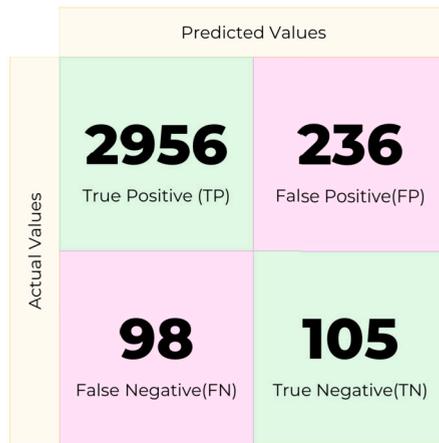

**Fig. 12.** Confusion matrix for AdaBoost.

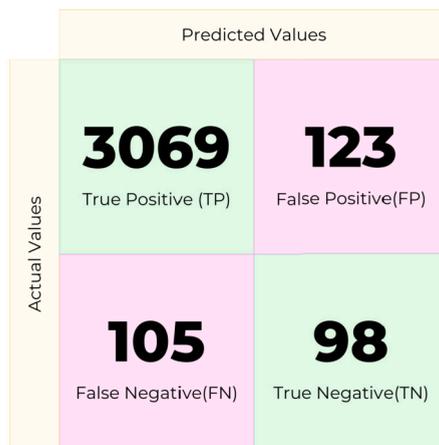

**Fig. 13.** Confusion matrix for XGBoost.

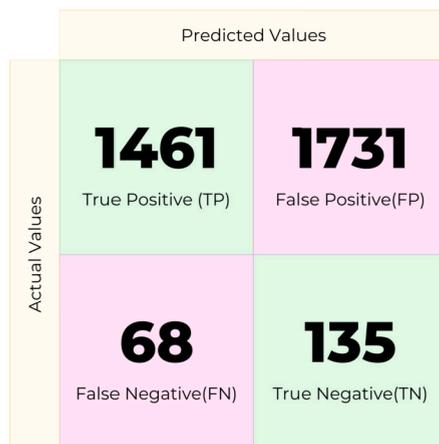

**Fig. 14.** Confusion matrix for SVC.

$$\mathbf{w} \cdot \mathbf{x} + b = 0 \tag{6}$$

### 3.5.7. Light gradient boosting

Light Gradient Boosting (LightGBM) is a gradient boosting algorithm for regression and classification applications that is fast and memory-efficient. It creates an ensemble of decision trees. LightGBM is optimized for large-scale datasets, focusing on performance and memory efficiency. In this experiment, the machine learning model created with this machine learning algorithm produced the confusion matrix in Fig. 15.

**Fig. 15.** Confusion matrix for Light Gradient Boosting.

**Fig. 16.** Confusion matrix for Naive Bayes.

*3.5.8. Naive Bayes*

Naive Bayes is a simple yet efficient classification algorithm based on Bayes' theorem. It assumes that features are independent of one another and compute probabilities to allocate data points to classes (eq (7)). The confusion matrix in Fig. 16 was obtained using Naive Bayes.

$$P(C|X) = \frac{P(C) \cdot \prod_{i=1}^{n} P(x_i|C)}{P(X)} \quad (7)$$

*3.5.9. Logistic regression*

Logistic Regression is a binary classification algorithm. It predicts probabilities by applying a sigmoid function (eq (8)) to input information. In this experiment, the machine learning model created with this machine learning algorithm produced the confusion matrix in Fig. 17.

$$P(Y=1|X) = \frac{1}{1 + e^{-(\beta_0 + \beta_1 X_1 + \beta_2 X_2 + \ldots + \beta_p X_p)}} \quad (8)$$

*3.6. Hyperparameter tuning*

Hyperparameter tuning can improve the performance and generalization of a machine learning model [42]. For each of the models used in our experiments, rigorous hyper-parameter optimization was performed using Grid Search CV and Randomized Search CV to get the best-performing models. Grid search CV works by creating a grid of all possible combinations of the hyperparameters and then evaluating the model for each combination using cross-validation. In contrast, a Randomized search CV works by randomly sampling a fixed number of combinations from the hyperparameters and then evaluating the model for each combination using cross-validation. The average score of the cross-validations is used to measure the model's performance and generate optimal hyperparameters that can be used. However, Grid search is computationally expensive since it explores the entire hyperparameter space. On the contrary, a Randomized search CV explores a subset of the space and tries to estimate the hyperparameters closest to the best ones. For certain models like LightGBM, Random Forest, and XGBoost, the types of hyperparameters and their number of combinations are abundant. As a result, a Randomized search CV was used to get the best hyperparameters for these models. On the other hand, models like Gradient Boosting, AdaBoost, and Logistic Regression require less exploration of the hyperparameters, which led us to use Grid search CV for these models. There was a drastic improvement in the performance of the models after using hyperparameter optimization compared to default hyperparameters. The best hyperparameters for the different models are displayed in Table 3. For the rest of the models, we found that the models performed best with the default parameters.

**Fig. 17.** Confusion matrix for Logistic Regression.

### 3.7. Ensemble machine learning model construction

Recent studies have shown that tree-based models and ensemble methods such as Random Forest and boosting algorithms can perform very well on tabular data, with problems that have clear decision boundaries and heterogeneous feature spaces [43,44]. This robust performance is a result of using decision trees as weak learners and reducing variance (in the case of random forest) and bias (in the case of boosting models) by combining multiple models. These models can outperform deep learning-based models by a great margin regarding tabular data. Driven by these recent findings and emphasis on ensemble methods, in this research, several ensemble models were constructed to predict life satisfaction in people. The ensemble approach combines several models to improve the overall performance and reliability of predictions. The created ensemble model incorporated a variety of classifiers, with a focus on boosting models like XGBoost, LGBoost, Gradient Boosting, and AdaBoost.

Several combinations of the models mentioned in the previous section were tested, and through empirical experimentation, it was found that the combination of `RF`, `GB`, and `LGB` yielded the best performance. The process is illustrated in Fig. 18. In addition, hyperparameter optimization was used to determine the class weights that give the best performance, and it was found that a class weight of {0: 5, 1: 0.09} for Random Forest gave the optimal outputs.

### 3.8. Large language models

#### 3.8.1. BERT

BERT, or Bidirectional Encoder Representations from Transformers, is a transformer-based machine learning architecture for natural language processing (NLP) pre-training [45]. Developed by Google, BERT's architecture consists of a multi-layer bidirectional Transformer encoder. For the BERT-Base model, there are 12 layers (transformer blocks), 768 hidden units, 12 self-attention heads, and a total of 110 million parameters. BERT is pre-trained on the BooksCorpus and English Wikipedia, which contain a combined total of 3.3 billion words.

In this study, BERT was chosen for life satisfaction prediction due to its deep bidirectional nature, allowing it to understand the context of words based on their surrounding text. It is particularly beneficial for interpreting the information embedded in texts. BERT currently holds state-of-the-art for several NLP tasks including question answering, natural language inference, text prediction, named entity recognition, sentiment analysis, language translation, text summarization, and text classification.

#### 3.8.2. BioBERT

BioBERT extends BERT's capabilities to the biomedical domain by pre-training on large-scale biomedical corpora [46]. It shares the same architecture as BERT but is further trained on biomedical texts such as PubMed abstracts and PMC full-text articles, enabling it to capture biomedical semantics more effectively.

**Table 3**
Best hyperparameters found using hyperparameter tuning on various models used in this research.

| Model | Hyperparameters |
|---|---|
| LightGBM (LGB) | boosting_type: gbdt<br>objective: binary<br>metric: binary_logloss<br>num_leaves: 31<br>learning_rate: 0.05<br>feature_fraction: 0.9 |
| Random Forest (RF) | n_estimators: 600<br>min_samples_split: 2<br>min_samples_leaf: 1<br>max_features: log2<br>max_depth: 780<br>criterion: gini |
| Gradient Boosting (GB) | n_estimators: 500<br>learning_rate: 1<br>max_depth: 1 |
| AdaBoost | base_estimator: clf<br>n_estimators: 600<br>random_state: 21<br>learning_rate: 1 |
| Logistic Regression | solver: liblinear<br>penalty: l2 |
| XGBoost | learning_rate: [0.05, 0.10, ..., 0.30]<br>max_depth: [3, 4, ..., 15]<br>min_child_weight: [1, 3, ..., 7]<br>gamma: [0.0, 0.1, ..., 0.4]<br>colsample_bytree: [0.3, 0.4, ..., 0.7]<br>booster: [gbtree, gblinear, dart] |

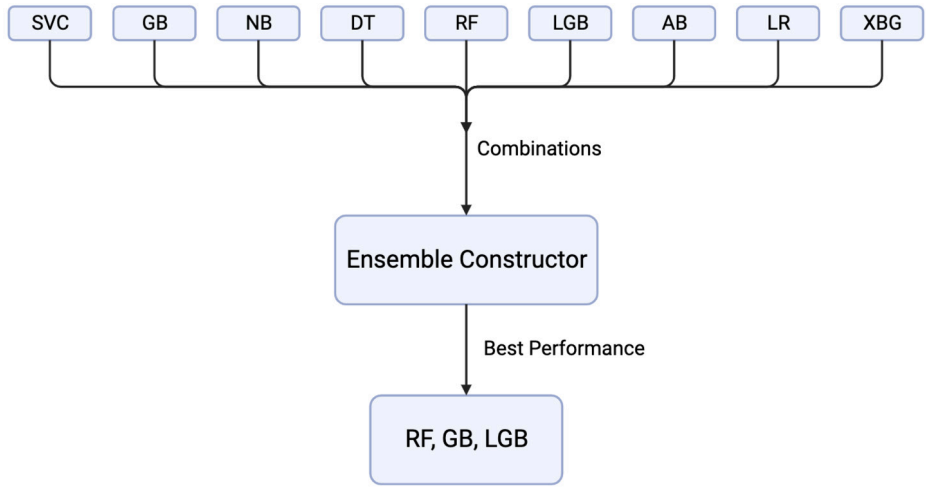

**Fig. 18.** The technique employed to get the best-performing ensemble model from a combination of all the models used.

The use of BioBERT is justified for life satisfaction prediction tasks that may involve biomedical terminology, especially when analyzing physical and mental health-related factors that significantly impact life satisfaction (Sec. 3.4.8).

### 3.8.3. ClinicalBERT

ClinicalBERT adapts BERT to the clinical domain by continuing the pre-training on clinical notes [47]. It utilizes the same transformer-based architecture as BERT but is fine-tuned on datasets comprising electronic health records (EHRs), which enables it to grasp medical jargon and patient narratives.

ClinicalBERT is pertinent for the classification task as the life satisfaction assessment involves physical and mental health-related data, and since it can better understand and process such specialized language.

*3.8.4. COReBERT*

CORe, or Clinical Outcome Representations, is a model designed to capture representations of clinical and biomedical outcomes [48]. It is initialized with BERT-Base and fine-tuned on 20.6B words from 2.1M clinical notes and 4.5M biomedical abstracts.

CORe could be advantageous for predicting life satisfaction when the outcomes are closely tied to clinical and biomedical events or patient health trajectories.

*3.9. Implementation details for LLM experiments*

All the experiments were run on the online platform Kaggle, with an NVIDIA P100 15 GB Tensor Core GPU having 29 GB of RAM. PyTorch [49] was used as the framework to conduct the experiments and to load the pre-trained base models from Hugging Face [50] using the Transformers [51] library. The WordPiece [52] tokenizers available for each model were used to tokenize the texts for processing by the language models and use the first 512 tokens due to the input limitation of BERT-like models. Each experiment was repeated five times and the mean with the standard deviation was reported in our results. All the models were trained for 200 epochs with an early stopping method. To avoid overfitting and boost model generalization on unobserved data, an early stopping strategy was used, keeping the macro averaged F1-score% as the monitor since we are working with imbalanced data. Furthermore, the AdamW [53] optimizer was used with a batch size of 16, and thorough hyperparameter tuning was conducted to find the best value of all the following: gradient accumulation steps = 10, learning rate of $1 \times 10^{-5}$, weight decay factor = 0.01, linear warm-up learning rate scheduler steps = 50, minimum delta for early stopping = 0.0001, and patience for early stopping = 10 epochs.

For the hyperparameter tuning, randomized search CV [54] was used, and the following ranges were defined for each hyperparameter: gradient accumulation steps $\in \{5, 10, 15, 20\}$, learning rate $\in [10^{-6}, 10^{-2}]$, weight decay factor $\in [0, 0.1]$, linear warm-up learning rate scheduler steps $\in \{10, 20, 50, 100\}$, minimum delta for early stopping $\in [10^{-4}, 10^{-2}]$, and patience for early stopping $\in \{1, 2, 3, 4, 5, 6, 7, 8, 9, 10\}$. 10 random combinations of hyperparameters were sampled from these ranges, and evaluated on a validation set by dividing the training set in a ratio of 80:20. The best hyperparameter combination was selected based on the highest macro F1-score% on the validation set.

*3.10. Performance metrics*

The machine learning models used in this study were evaluated using the performance metrics precision, recall, accuracy, F1-score, and AUC ROC.

**Precision:** Positive Predictive Value or Precision, as shown in eq (9), is the ratio of correctly predicted positive samples to the total number of predicted positive values.

$$Precision = \frac{TP}{TP + FP} \quad (9)$$

**Recall:** Sensitivity or Recall is calculated as the ratio between the number of positive samples to the total number of positive samples. The high recall represents more positive samples. It helps to detect positive samples, and is given in eq (10) as:

$$Recall = \frac{TP}{TP + FN} \quad (10)$$

**F1-score:** F1-score is considered a more reliable performance metric. It represents the combination of precision and recall, which gives a score between 0 and 1. It is represented in eq (11) as:

$$F1 - score = \frac{2 \times TP}{2 \times TP + FP + FN} \quad (11)$$
$$= 2 \times \frac{Precision \times Recall}{Precision + Recall}$$

**Accuracy:** Eq (12) shows the formula for accuracy with respect to True Positives (TP), True Negatives (TN), False Positives (FP), and False Negatives (FN):

$$Accuracy = \frac{TP + TN}{TP + TN + FP + FN} \quad (12)$$

**ROC:** ROC or Receiver Operator Characteristics curve is a performance metric graph that plots the TP rate (Sensitivity) against the FP rate (Specificity), shown respectively in eq (13) and eq (14), where,

$$TPR = \frac{TP}{TP + FN} \quad (13)$$

$$FPR = \frac{FP}{FP + TN} \quad (14)$$

AUC, or Area Under The Curve, provides an aggregated measure at all classification thresholds. This study uses the AUC ROC score as an evaluation metric. It is plotted as subplots for DecisionTreeClassifier, Random Forest, Gradient Boosting, AdaBoost, and XGBoost, and scores are obtained, respectively.

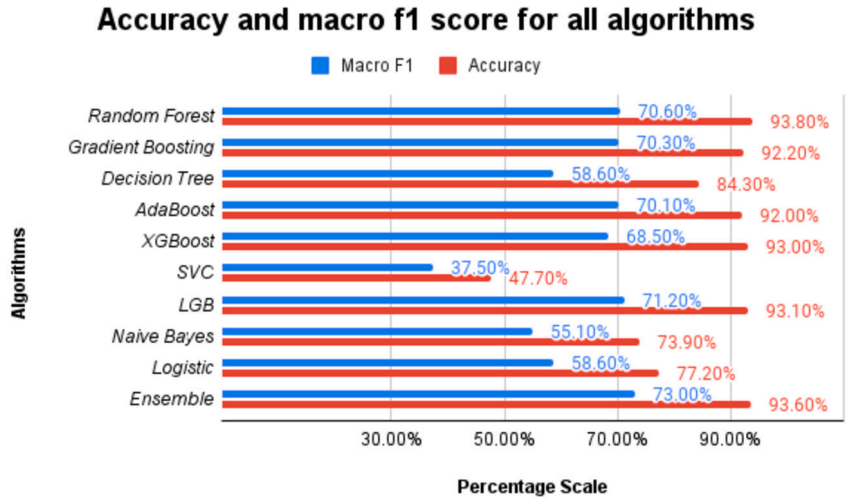

**Fig. 19.** Comparison between accuracy and macro F1 score for different models.

**Table 4**
Comparative evaluation of all the machine learning models used in this study. The mean of five experiments with different random seeds is reported with the standard deviation to provide a statistically significant representation of the results.

| Model | Accuracy (%) | F1 (%) | Precision (%) | Recall (%) | ROC (%) |
| --- | --- | --- | --- | --- | --- |
| Random Forest | **93.80 ± 0.1000** | 70.60 ± 0.5000 | **72.00 ± 0.6000** | 69.30 ± 0.4000 | 89.11 ± 0.1000 |
| Gradient Boosting | 92.20 ± 0.0000 | 70.30 ± 0.0000 | 67.90 ± 0.0000 | 73.70 ± 0.0000 | 88.40 ± 0.0000 |
| Decision Tree | 84.30 ± 0.4000 | 58.60 ± 0.2000 | 57.10 ± 0.2000 | 66.40 ± 0.4000 | 66.40 ± 0.4000 |
| AdaBoost | 92.00 ± 0.0000 | 70.10 ± 0.0000 | 67.50 ± 0.0000 | 74.30 ± 0.0000 | 86.60 ± 0.0000 |
| XGBoost | 93.00 ± 0.2000 | 68.50 ± 0.6000 | 68.70 ± 0.9000 | 68.20 ± 0.5000 | 88.00 ± 0.4000 |
| SVC | 47.70 ± 0.0000 | 37.50 ± 0.0000 | 51.40 ± 0.0000 | 56.10 ± 0.0000 | 60.70 ± 0.0000 |
| LGB | 93.10 ± 0.1000 | 71.20 ± 0.2000 | 70.00 ± 0.3000 | 72.70 ± 0.6000 | 89.10 ± 0.1000 |
| Naive Bayes | 73.90 ± 0.0000 | 55.10 ± 0.0000 | 56.90 ± 0.0000 | 75.30 ± 0.0000 | 83.00 ± 0.0000 |
| Logistic Regression | 77.20 ± 0.0000 | 58.60 ± 0.0000 | 58.80 ± 0.0000 | **80.70 ± 0.0000** | 88.80 ± 0.0000 |
| Ensemble | 93.60 ± 0.1000 | **73.00 ± 0.400** | 71.90 ± 0.4000 | 74.30 ± 0.5000 | **89.70 ± 0.0000** |

## 4. Performance and evaluation

### 4.1. Performance on machine learning algorithms

The accuracy of the predictions made by the different machine learning models in this study has been measured using five performance metrics, namely, precision (eq (9)), recall (eq (10)), F1-score (eq (11)) accuracy (eq (12)) and ROC score. Using the confusion matrix of all the models, we calculated the performance metrics, which have been listed in Table 4. To better represent the performance of our models, the comparison of macro F1 and accuracy scores were shown in Fig. 19. The F1-score is a harmonic mean of precision and recall, and it provides a balanced measure of the classifier's performance for imbalanced datasets. Accuracy, on the other hand, is the ratio of correctly predicted samples to the total number of samples. In conclusion, the ensemble classifier beat all other models regarding F1-score and precision, which are important in case of imbalanced data such as ours. More detailed analysis and discussion of the results are presented in Sec. 5.1.

### 4.2. Performance on large language models

We present the results of the experiments performed with the LLMs using the generated sentences from Sec. 3.4.9 in Table 5. The table shows the accuracy, macro-averaged f1-score, precision, recall, and ROC of the binary text classification performed on the sentences in percentage. Detailed analysis and discussion of the results are presented in Sec. 5.2.

### 4.3. Statistical significance of model performance

To assess the statistical significance of the differences in performance between the machine learning models and the large language models, paired t-tests were conducted for the Accuracy, F1-score, Precision, Recall, and ROC metrics using the SciPy library [55]. The p-values from the paired t-tests are reported in Table 6.

For the machine learning models, the results show that the Random Forest model performed significantly better than the Gradient Boosting model in terms of Accuracy, Precision, Recall, and ROC (p < 0.05). However, the differences in F1-score were not statistically significant. Additionally, the differences in performance between the LGB, XGBoost, and Ensemble models were not statistically

**Table 5**

Model performance and evaluation results on the large language models, represented in percentage. The best results are marked in bold. The mean of five experiments with different random seeds is reported with the standard deviation to provide a statistically significant representation of the results.

| Model | Accuracy (%) | F1 (%) | Precision (%) | Recall (%) | ROC (%) |
| --- | --- | --- | --- | --- | --- |
| BERT | 93.58 ± 0.0012 | 71.45 ± 0.0022 | 71.45 ± 0.0011 | 71.45 ± 0.0032 | 89.06 ± 0.0101 |
| BioBERT | 93.53 ± 0.0019 | **73.21 ± 0.0034** | 71.81 ± 0.0036 | **74.87 ± 0.0023** | **89.12 ± 0.0015** |
| ClinicalBERT | **93.74 ± 0.0021** | 71.24 ± 0.0032 | **71.92 ± 0.0023** | 70.63 ± 0.0013 | 88.72 ± 0.0027 |
| COReBERT | 93.56 ± 0.0012 | 70.98 ± 0.0034 | 71.23 ± 0.0017 | 70.74 ± 0.0034 | 87.96 ± 0.0017 |

**Table 6**

P-values from the paired t-tests comparing the performance metrics of the machine learning models and large language models.

| Model Comparison | Accuracy | F1-score | Precision | Recall | ROC |
| --- | --- | --- | --- | --- | --- |
| Machine Learning Models | | | | | |
| Random Forest vs. Gradient Boosting | **0.001** | 0.678 | **0.033** | **0.019** | **0.002** |
| LGB vs. Ensemble | **0.002** | 0.116 | 0.234 | 0.123 | 0.112 |
| XGBoost vs. Ensemble | **0.003** | 0.274 | 0.287 | 0.089 | 0.090 |
| Large Language Models | | | | | |
| BioBERT vs. BERT | 0.270 | **0.003** | 0.277 | **0.001** | 0.223 |
| ClinicalBERT vs. BioBERT | **0.015** | 0.071 | **0.045** | **0.001** | 0.087 |
| COReBERT vs. ClinicalBERT | 0.053 | 0.123 | 0.072 | 0.372 | 0.051 |

significant for most of the metrics, except for Accuracy, where the Ensemble model outperformed the LGB and XGBoost models (p < 0.05).

For the large language models, the results indicate that BioBERT significantly outperformed BERT in terms of F1-score and Recall (p < 0.05), while ClinicalBERT had significantly higher Accuracy and Precision compared to BioBERT (p < 0.05). The differences in performance between ClinicalBERT and COReBERT were not statistically significant for most of the metrics.

These findings suggest that the Random Forest model and the BioBERT model are the top-performing models in the machine learning and large language model categories, respectively, for the task of life satisfaction prediction. The superior performance of these models highlights their suitability and effectiveness in capturing the complex and multifaceted determinants of subjective well-being.

*4.4. Error analysis*

This section provides an in-depth error analysis of the machine-learning models used in this study. Fig. 20 represents a stacked bar plot with the false positives and false negatives of the different models used in our research. The analysis aims to provide insights into the performance and characteristics of these models, shedding light on their strengths and weaknesses. The observations made here contribute to a better understanding of the predictive capabilities of the models, enhancing the overall validity and reliability of our research findings.

The stacked bar plot demonstrates that all models exhibit more false positives than false negatives. This observation indicates a tendency for the models to erroneously predict positive instances of life satisfaction when they are, in fact, negative. While false positives can be undesirable, exploring the extent of this imbalance and its implications is crucial. Further examination of the stacked bar plot reveals that some models (LGB, RF, GB, AB, XGB, Ensemble) demonstrate a nearly equal number of false positives and false negatives. This suggests a certain level of consistency in their predictive performance, as they display comparable errors in both directions. On the other hand, one specific model stands out with significantly higher occurrences of false positives and false negatives. This model requires closer scrutiny to identify potential sources of bias or limitations in its design. A notable distinction between the boosting algorithms and the other models regarding the structure of false positives and false negatives can be seen. The boosting algorithms exhibit a symmetrical pattern, indicating a relatively balanced misclassification behavior. In contrast, the remaining models display an asymmetric structure, implying a varying degree of bias towards either false positives or false negatives. This discrepancy could indicate inherent differences in the learning and decision-making mechanisms employed by these algorithms. The error analysis based on the stacked bar plot offers valuable insights into the performance characteristics of the models utilized in our research. The prevalence of false positives across all models warrants further investigation to determine potential factors contributing to this trend. Understanding the reasons behind such misclassifications can facilitate improvements in future iterations of the models, leading to more accurate predictions and enhanced practical utility. Acknowledging that the error analysis presented here is based on a specific dataset and modeling approach is essential. The generalizability of the findings to other datasets or model architectures may vary. Nonetheless, these observations contribute to the broader body of knowledge in the field of machine learning, and the observed trends, including the prevalence of false positives and the varying characteristics between models, provide valuable insights for understanding and improving the performance of predictive models in the domain of life satisfaction prediction.

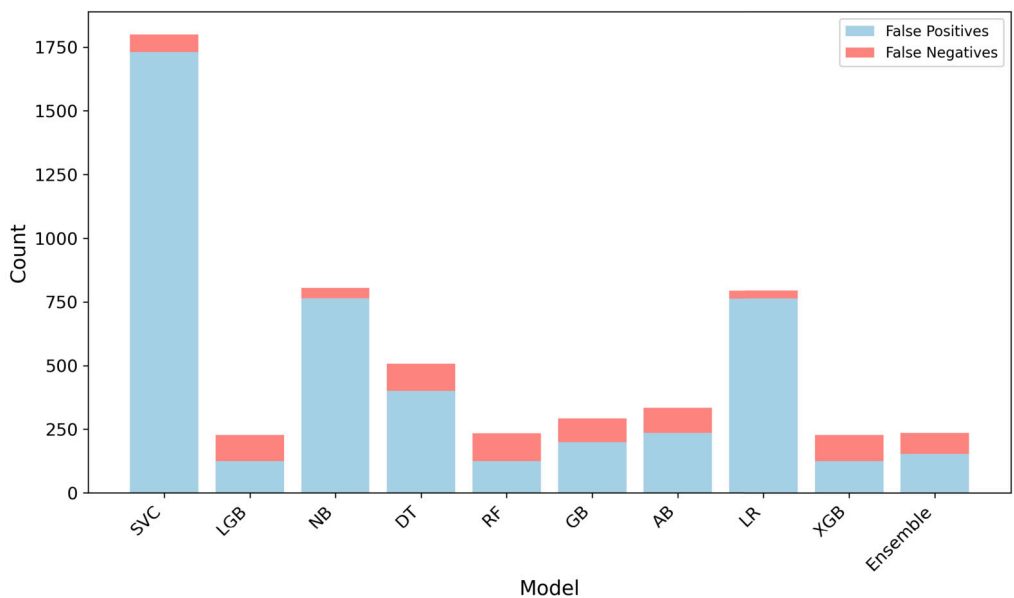

**Fig. 20.** Error analysis of various machine learning models for life satisfaction.

**Table 7**
Ablation results of the machine learning models comparing different data resampling techniques. The best results are shown in bold.

|  | Without Resampling | | Oversampling Only | | Undersampling Only | | Over & under sampling | |
| --- | --- | --- | --- | --- | --- | --- | --- | --- |
| Model | Acc.(%) | F1(%) | Acc.(%) | F1(%) | Acc.(%) | F1(%) | Acc.(%) | F1(%) |
| Random Forest | 89.20 | 68.30 | 94.10 | 71.80 | 75.90 | 57.60 | 93.80 | 70.60 |
| Gradient Boosting | 94.60 | 67.40 | 94.30 | 65.90 | 76.20 | 57.60 | 92.20 | 70.30 |
| Decision Tree | 88.20 | 58.90 | 89.00 | 60.80 | 69.60 | 52.10 | 84.30 | 58.60 |
| AdaBoost | 93.40 | 68.60 | 94.00 | 71.80 | 75.00 | 57.10 | 92.00 | 70.10 |
| XGBoost | 94.40 | 62.80 | 94.60 | 66.00 | 76.00 | 57.30 | 93.00 | 68.50 |
| SVC | 94.10 | 48.50 | 83.40 | 54.30 | 58.40 | 43.80 | 47.70 | 37.50 |
| LGB | 93.90 | 69.70 | 94.00 | 70.50 | 76.10 | 58.00 | 93.10 | 71.20 |
| Naive Bayes | 79.20 | 59.40 | 79.40 | 59.40 | 68.40 | 52.50 | 73.90 | 55.10 |
| Logistic Regression | 94.40 | 72.60 | 85.60 | 65.20 | 75.00 | 57.10 | 77.20 | 58.60 |
| Ensemble Classifiers | **94.80** | 68.80 | 94.70 | 70.02 | 75.30 | 57.20 | 93.60 | **73.00** |

### 4.5. Ablation studies

In this section, we perform ablation studies to examine the effect and contribution of the following factors on the overall performance of the machine learning models: (i) the applied data resampling techniques and (ii) the feature selection methods.

#### 4.5.1. Ablation on data resampling

In Sec. 3.4.6, a dual balancing strategy was applied to resample the data by applying both oversampling and undersampling to the original data distribution. Table 7 shows the results of the models without applying any data resampling technique, applying only oversampling and only undersampling, respectively.

The ablation experiments reveal significant insights into the impact of resampling strategies on model performance. Without resampling, most models exhibit high accuracy, yet the F1 scores are not optimal, indicating a potential imbalance in the precision-recall trade-off. Oversampling alone appears to improve the F1 scores for certain models like SVC and Gradient Boosting, suggesting that increasing the representation of minority classes aids in achieving a more balanced classification. However, this comes at the cost of reduced accuracy, which may be attributed to the models' overfitting of the oversampled data.

Undersampling solely shows a marked decrease in both accuracy and F1 scores across all models. This decline indicates that while reducing the majority class can help mitigate the class imbalance, it may also lead to the loss of valuable information, thereby impairing the model's ability to generalize. The combined approach of oversampling and undersampling presents a mixed outcome. For instance, the Ensemble Classifiers demonstrate an improvement in F1 score, achieving the best result among all resampling techniques, which underscores the efficacy of a dual balancing strategy in managing class distribution without significantly compromising the accuracy.

These observations highlight the necessity of a balanced approach to data resampling in predicting life satisfaction from the dataset at hand. It is imperative to consider the trade-offs between accuracy and F1 score and to select a resampling strategy that aligns with the specific requirements of the application domain. Furthermore, the results advocate for the exploration of hybrid resampling methods that can harness the strengths of both oversampling and undersampling to optimize model performance, which was ultimately adopted in this study.

**Table 8**
Ablation results of the machine learning models comparing different feature elimination and reduction techniques. The best results are shown in bold.

| Model | Without RFECV | | PCA (95%) | | PCA (90%) | | With RFECV | |
|---|---|---|---|---|---|---|---|---|
| | Acc.(%) | F1(%) | Acc.(%) | F1(%) | Acc.(%) | F1(%) | Acc.(%) | F1(%) |
| Random Forest | 94.30 | 70.30 | 89.10 | 56.00 | 88.50 | 57.40 | **93.80** | 70.60 |
| Gradient Boosting | 93.40 | 72.50 | 84.10 | 59.30 | 84.20 | 59.50 | 92.20 | 70.30 |
| Decision Tree | 84.10 | 57.60 | 78.70 | 52.10 | 78.70 | 53.20 | 84.30 | 58.60 |
| AdaBoost | 90.50 | 67.80 | 83.00 | 55.60 | 83.20 | 55.10 | 92.00 | 70.10 |
| XGBoost | 93.40 | 69.90 | 89.70 | 55.60 | 88.80 | 56.50 | 93.00 | 68.50 |
| SVC | 49.20 | 38.50 | 88.10 | 55.60 | 88.00 | 57.00 | 47.70 | 37.50 |
| LGB | 91.40 | 69.40 | 83.00 | 57.10 | 83.40 | 55.50 | 93.10 | 71.20 |
| Naive Bayes | 82.60 | 59.60 | 82.20 | 54.90 | 80.70 | 55.80 | 73.90 | 55.10 |
| Logistic Regression | 79.50 | 60.00 | 77.20 | 58.50 | 77.10 | 58.40 | 77.20 | 58.60 |
| Ensemble | 93.20 | 71.90 | 81.60 | 59.80 | 81.80 | 60.20 | 93.60 | **73.00** |

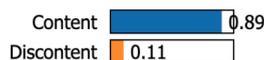

**Fig. 21.** Case-1 Prediction probability shows high probability for 'Content' class.

*4.5.2. Ablation on feature selection techniques*

In our experimental pipeline, several feature selection and feature elimination techniques were employed to extract and find the most important determinants for predicting life satisfaction. These techniques include removing features having zero variance and applying recursive feature elimination using cross-validation (RFECV). However, it is crucial to examine how well the models perform without applying any feature elimination technique in order to understand the contribution of these methods to the overall model performance and to justify the robustness and strength of our approach. Moreover, it is important to explore other feature reduction techniques, such as principal component analysis (PCA), to show that the methods that were used gave the best possible results.

To this goal, three ablation experiments were performed with all the machine learning models: (i) Without applying RFECV. (ii) PCA (with 95% variance) instead of RFECV, and (iii) PCA (with 90% variance) instead of RFECV. Table 8 shows the results of these experiments, compared to the original results with RFECV. After data preprocessing steps, without applying RFECV, the number of features found was 100. By applying PCA with 95% and 90% variance, the number of features found were 24 and 22, respectively, which is close to the number of features found using RFECV, which was 27.

Without applying any feature elimination, the models exhibit a significant drop in performance, particularly in terms of F1 score. For instance, the Ensemble Classifiers model's F1 score decreases from 73.00% to 71.90% when RFECV is not used. This highlights the importance of feature selection in enhancing the models' ability to focus on the most salient determinants of life satisfaction, thereby improving their overall predictive accuracy and generalization capabilities.

Furthermore, the experiments with PCA-based feature reduction demonstrate that the RFECV approach outperforms these alternative techniques. Even with PCA retaining 95% of the variance, the model performance lags behind the RFECV-based results. This suggests that the feature selection process employed in this study, which leverages the inherent relationships between the features and the target variable, is more effective in identifying the critical predictors of life satisfaction than the more generic dimensionality reduction offered by PCA. Thus, our streamlined feature set using RFECV not only enhances the computational efficiency of the models but also promotes interpretability, a crucial aspect for real-world applications in the mental health domain.

*4.6. Explainable AI*

Although decisions made by a machine learning model might be very accurate, they often come with a question asking how the model made this particular decision. The study uses explainable AI to get explanations for the decisions made by the machine learning model. The decision-making process for different cases has been explained in the sample investigations 4.6.1 and 4.6.2, along with their prediction probabilities.

*4.6.1. Case-1*

The machine learning model predicts with a probability of 89% that the person is content with his life, as illustrated in Fig. 21.

The answers made by the subject have associated values that either reward or penalize the probability of them being identified as content/discontent depending on whether these values fall within the positive thresholds learned by the machine learning model.

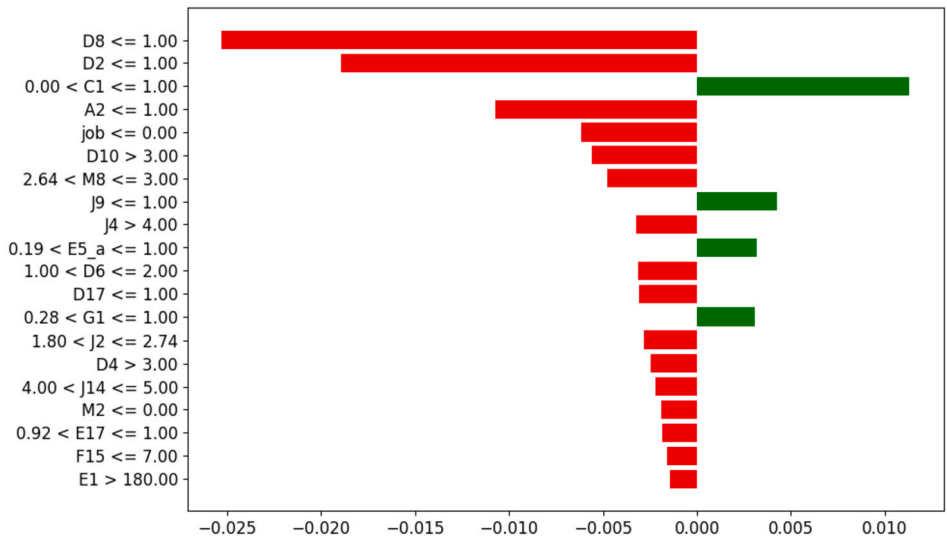

**Fig. 22.** Case-1 explanation threshold. The majority of decision scores are negative for the 'Content' class.

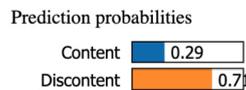

**Fig. 23.** Case-2 Prediction probability shows high probability for the 'Discontent' class.

Fig. 22 shows the thresholds along with the rewards as green bars and penalties as red bars in calculating the sample's class. For example, the person in observation reports his health condition as 'Average' ($A2 <= 1.0$) and that he finds himself rarely worrying ($D8 <= 1.0$). These two values lie in the threshold that penalizes the 'discontent' class. However, he also reports that he had rarely been to cinemas, theaters, or any similar form of amusement ($J9 <= 1.0$), which is the threshold that rewards the 'discontent' class. Similarly, the answers to the remaining questions provided by the subject either reward or penalize the total score. The sum of all these rewards and penalties is taken as the total score. Here, the score is highly negative for the discontent class, so the subject is predicted to be content in life. It can be seen from this example how Explainable AI justifies the decisions made by the machine learning model and, in doing so, makes our approach trustworthy and reliable.

*4.6.2. Case-2*

The machine learning model predicts with a probability of 71% that the person in case-1 is Discontent with his life, as illustrated in Fig. 23.

Similar to case-1, the answers made by the subject have associated values that either reward or penalize the probability of her being identified as content/discontent. Fig. 24 shows the thresholds along with the rewards as green bars and penalties as red bars in calculating which class the sample belongs to. The person in observation reports that she sometimes finds herself depressed ($D2 > 1.6$) and that she often worries ($D8 > 2.5$), which lies in the threshold that rewards the discontent class. Nevertheless, she also reports that she primarily talks to her spouse about personal and serious problems ($E17 > 1.19$), which is the threshold that penalizes the Discontent class. Likewise, the sum of the rest of the rewards and penalties for the answers to the remaining questions provided by the subject show that the score is highly positive for the discontent class, so the subject is predicted to be discontent in life.

*4.7. Age group insights*

One can estimate that the primary determinants of life satisfaction are not the same across different age brackets. So, the primary determinants across four different age brackets were explored, and it was compared how they change throughout one's lifetime. The selected age brackets are as follows: young age (ages 16-21), early adulthood (ages 22–34), middle age (ages 35–44), old age (ages 45–64) [56]. Fig. 25 compares these primary determinants we have extracted for the four different age brackets. These were extracted using RFECV feature importance analysis for the different age brackets. In this figure, the four radar plots represent the importance of the top five features extracted within the age brackets. It can be observed that for every age bracket, health is the most important determinant of life satisfaction. For the younger population (16-34), depression is the next most important determinant, while for those in their later adulthood, worrying is the next primary concern. The radar plots further show that in the case of the population in the age range 16-21, tension, mood, and worry play very important roles in determining life satisfaction. During these periods, people usually go through major physical and mental changes, which explains why mood is an important factor. Moreover, during this period, they step on to adulthood, passing from teenagehood, learning to take responsibility, and thus, worrying about life and reality. When reality hits hard, it is incumbent that they go through mental pressure and tension, which is also observed in the plots. Furthermore, in a slightly older population ranging from age 22 to 34, the primary sources of income, employment status, and emotional stability play crucial roles in determining their state of contentment. During this period, a person starts building their own life, perhaps even a family. As a result, the ability to provide for themselves or their family plays a massive role in determining their life satisfaction. This is greatly affected by their employment status since unemployment leads to smaller or no sources of income. Lastly, all these changes naturally take a toll on a person's emotional stability, another important factor in determining the person's life contentment, as depicted in the radar plots. Another interesting observation is that the top five determinants for the two age groups, middle age and old age, are the same, including health, worrying, depression, tension, and long-term health problems. This suggests that physical and mental health dominantly dictate life satisfaction for those above 35. However, it can be seen that physical health is more important in middle age, while mental health is more important in old age.

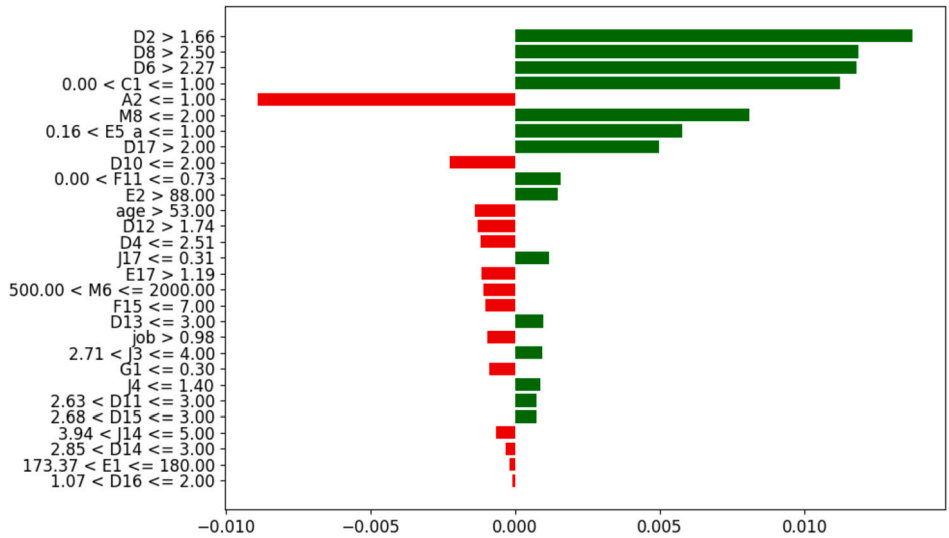

**Fig. 24.** Case-2 explanation threshold. The majority of decision scores are positive for the 'Discontent' class.

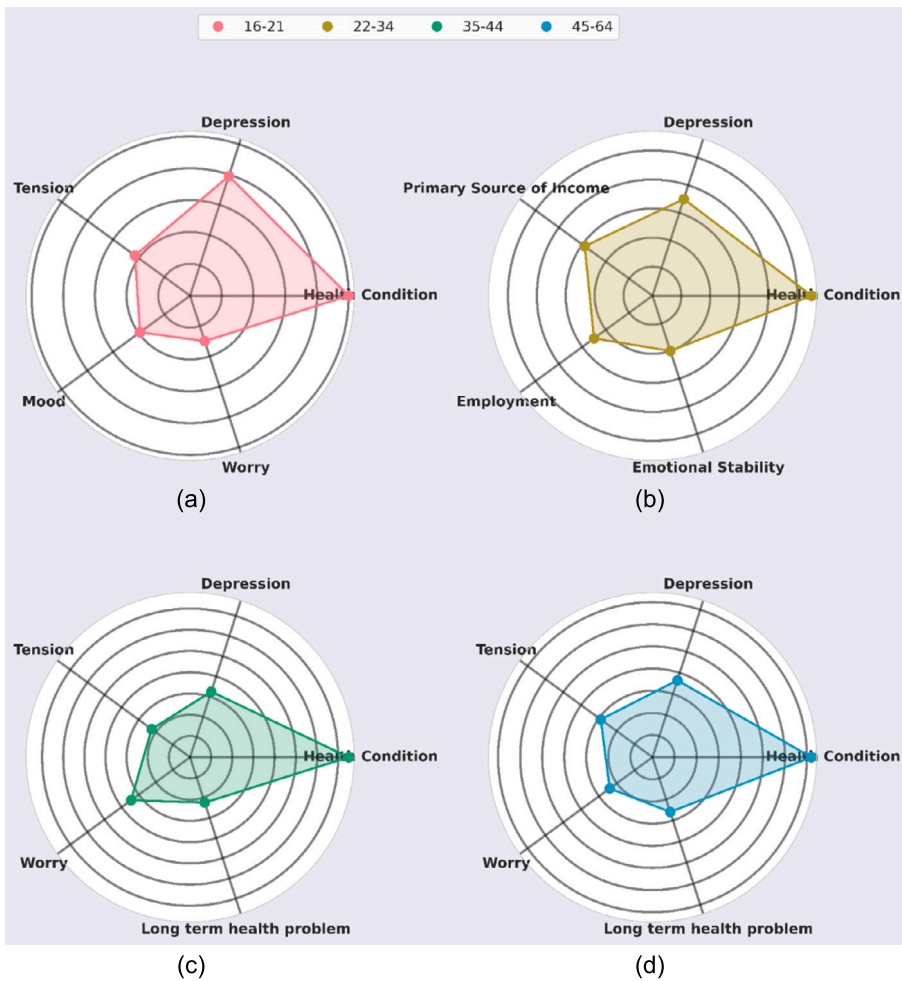

**Fig. 25.** Radar plots showcasing the importance of the top five features extracted for each age bracket. The more a point is towards a feature, the greater its importance, and vice-versa. Shown in (a), (b), (c), and (d) are the plots representing the age ranges of 16 to 21, 22 to 32, 35 to 44, and 45 to 64, respectively.

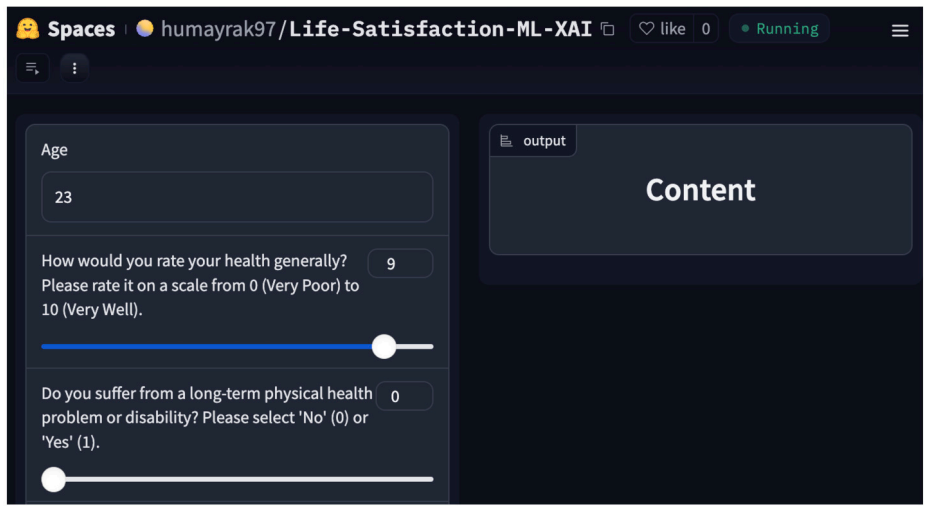

**Fig. 26.** Case-1 Prediction for 'Content' class.

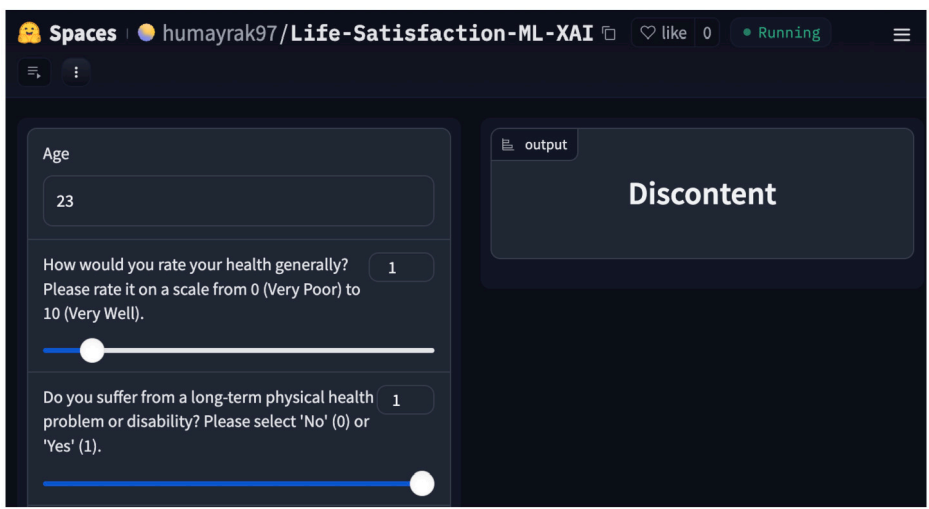

**Fig. 27.** Case-2 Prediction for 'Discontent' class.

*4.8. Deployment*

An interactive app was developed in this study to extend the reach and impact of the research on predicting life satisfaction using machine learning. The Gradio [57] powered application features a user-friendly interface for answering the questionnaire prepared in this study. It allows individuals to answer a carefully crafted questionnaire that aligns with the objectives of our research. Fig. 26 and 27 illustrate the outcomes of the application, showcasing the dichotomous classification of individuals into content and discontent states based on their responses. Users will get the prediction of the state of contentment in their lives. Behind the scenes, the machine learning models analyze these inputs and provide real-time predictions of life satisfaction. The app was hosted on a secure platform, ensuring data privacy, and we plan to offer continuous maintenance and updates. This deployment makes the research findings accessible to a broader audience and encourages the use of machine learning and Explainable AI to understand and enhance subjective well-being. The app can be found at: https://huggingface.co/spaces/humayrak97/Life-Satisfaction-ML-XAI.

## 5. Discussion

*5.1. Performance analysis of machine learning algorithms*

A closer look at Table 4 suggests that the deployed ensemble model performs best among all the machine learning models in terms of F1 score, giving 73.00%. This shows the superiority of the ensemble model over any other model since the dataset we used is imbalanced. The F1 score, a harmonic mean of precision and recall, reflects the model's balanced capability to navigate false positives and false negatives. Specifically, an elevated F1 score of 73.00%indicates that the model has effectively harmonized its predictive capacity between false positives (instances wrongly identified as positive) and false negatives (positive instances incorrectly deemed negative), which is of paramount importance in contexts where both type I and type II errors carry significant implications. Particularly for life satisfaction prediction, misclassifications (either in the form of false positives or false negatives) could potentially skew analyses or interventions based on these predictions. Hence, the necessity for a model that proficiently manages this trade-off is evident.

This study employs decision tree-based and boosting models, including gradient boosting, LGB, AdaBoost, and XGBoost, due to their notable advantages in predictive performance and robustness. Boosting models, in particular, demonstrate a significant performance advantage over other models, with gradient boosting leading the pack. These models optimize performance by sequentially combining weak learners, typically decision trees, to build a robust predictive model. This sequential process allows boosting algorithms to refine predictive accuracy by correcting errors made by previous models. Additionally, boosting algorithms adaptively assign more weight to misclassified instances in successive trees, enabling them to effectively capture intricate patterns and nuanced relationships within the data. This adaptability is crucial for predicting life satisfaction from complex, multifaceted datasets, as it allows the models to manage potential noise and complexities inherent in the data while maintaining refined predictive accuracy. On the other hand, the efficacy of the Random Forest model in our study can be attributed to its ensemble nature, which aggregates the outputs of multiple decision trees to mitigate the risk of overfitting and provide a stabilizing effect, especially beneficial for high-dimensional datasets. Furthermore, Random Forests exhibit resilience against overfitting, robustness to outliers, and non-linearities, owing to their randomized bootstrapping and feature selection techniques. These characteristics enable Random Forests to manage unbalanced datasets effectively while maintaining predictive accuracy and providing a degree of model interpretability. This is confirmed from the ablation study presented in Table 7, where the Random Forest model demonstrates a high accuracy and F1 score even without applying any data resampling technique, suggesting an intrinsic robustness to class imbalance.

Table 4 further reveals that the Support Vector Classifier (SVC) performs poorly with regard to our classification task at hand, compared to all other models. It exhibits an F1 score of only 37.50%. The nature of the classification task and the characteristics of the dataset can have a significant impact on the performance of different machine learning algorithms. In the context of this study, the prediction of life satisfaction as a binary outcome represents a challenging classification problem, particularly due to the inherent complexities and subtleties involved in quantifying subjective well-being. The SVC model, while generally effective in capturing non-linear relationships and handling high-dimensional data, may struggle when faced with imbalanced datasets [58], as is the case in this study. Life satisfaction, being a multifaceted construct, can exhibit a complex decision boundary that may not be easily captured by the SVC's hyperplane-based approach. Moreover, from Fig. 7, we identify that the extracted feature list used in our classification task contains five kinds of data (physical, mental, social, economic, and cultural). The SVC's sensitivity to outliers and its underlying assumptions of linearity and Gaussian distributions may not be well-suited to the heterogeneous nature of the features involved in predicting life satisfaction, resulting in lower performance of the model.

Lastly, by distilling our feature set down to 27 key questions from the initial 243, our approach ensured computational efficiency, minimized the risk of overfitting, and enhanced model interpretability. The experimental results of Table 7 reveal that the dimensionality reduction was accomplished without a notable compromise in predictive accuracy, affirming that the extracted features possessed a heightened signal about life satisfaction. Furthermore, our approach's merger of machine learning with explainable AI (XAI) anchored the model's predictive capacity within a framework of interpretability and trust. It crafted a bridge between high-dimensional data and actionable insights and ensured that the predictive decisions were transparent, thereby maintaining ethical and user-trust considerations. This reciprocity in comprehension is pivotal for real-world applicability, ensuring the translation of our findings into practical, ethical, and user-centered applications in the mental health domain.

*5.2. Performance analysis of LLMs*

The performance of the large language models (LLMs) in predicting life satisfaction, as presented in Table 5, provides valuable insights into the nature of the task and the suitability of different models for this purpose. Among the tested LLMs, BioBERT achieves the highest macro-averaged F1-score of 73.21%, indicating its superior ability to balance precision and recall in the binary classification of life satisfaction. This result suggests that the detection of life satisfaction, as expressed through the generated sentences, can be considered a task more closely aligned with the biomedical domain rather than the clinical domain represented by ClinicalBERT.

The strong performance of BioBERT can be attributed to the breadth and diversity of its training corpus, which includes a wide range of biomedical literature, including scientific publications, clinical notes, and health-related web content. This broad exposure to various aspects of human well-being, health, and lifestyle factors may have equipped BioBERT with a more comprehensive understanding of the multifaceted determinants of life satisfaction, ultimately leading to its superior predictive capabilities. In contrast, the relatively lower performance of ClinicalBERT and COReBERT suggests that their training corpora, which are more specialized in the clinical domain, may not capture the full scope of factors influencing life satisfaction. While these models excel in clinical tasks, their narrower focus may limit their ability to generalize and identify the social, psychological, and lifestyle-related aspects that contribute to an individual's overall life satisfaction.

The consistent performance across all LLMs, with accuracy around 93.5% and ROC around 88.5%, highlights the inherent suitability of these powerful language models for the task of life satisfaction prediction. However, the divergence in their F1 scores and precision-recall characteristics underscores the importance of model selection based on the specific requirements and priorities of the application domain. Thus, the results suggest that life satisfaction prediction is a diverse task that extends beyond the clinical domain, requiring a more comprehensive understanding of the biomedical and psychosocial factors contributing to an individual's well-being. The superior performance of BioBERT in this context emphasizes the value of leveraging language models trained on diverse corpora that capture the broader aspects of human life and experience.

The results of the large language models (LLMs) provide a valuable complement to the insights gained from the machine learning models in the previous subsection. While the machine learning models were trained on a structured dataset of features, the LLMs are able to leverage the rich information contained within natural language sentences to predict life satisfaction. This allows the LLMs to capture the complex, multifaceted aspects of life satisfaction that may not be fully represented in the structured data alone. By combining the strengths of both approaches, we gain a more comprehensive understanding of the factors influencing life satisfaction and the suitability of different model architectures for this prediction task.

*5.3. Feature analysis*

*5.3.1. Social and economic implications*

The LifeWell survey shown in Fig. 7 highlights key factors linked to life satisfaction in this study. This questionnaire helps individuals identify critical factors influencing their happiness. Furthermore, our machine learning approach can scale mental healthcare and support individuals who cannot afford a therapist [59]. A deeper dive into the radar charts of Fig. 25 reveals how significant social and economic indicators like employment status are in measuring the life satisfaction of people. This study finds that economic factors, such as job status, satisfaction (F15), and financial well-being (M2, M6, M8), significantly impact life satisfaction. Employment and job contentment offer purpose and stability, fostering social integration. Financial wellness, seen in income, healthcare spending, and self-rated financial status, enables resource access and meets basic needs, further influencing life satisfaction. Moreover, we have shown the direct impact of the social indicators on people's physical and mental health, which is also noticeable from the figure. Social factors, like support (E17), relationship status (G1), frequency of interaction with relatives (J2), and family visits (J17), significantly impact well-being. A strong social network and active engagement enhance life satisfaction, offering support, belonging, and social opportunities. Conversely, social isolation can decrease satisfaction. These findings have important social and economic implications for practice and policy, as they provide evidence-based recommendations for improving subjective well-being and quality of life.

*5.3.2. Implications of culture-related features*

Among the 273 features, 27 were identified as the most significant. Several significant elements include culture-related indicators that reflect cultural values, such as the frequency of traveling abroad for holidays or family visits (J17), attending film, concert, or theater events (J9), and reading newspapers (J14), which have a critical impact on life satisfaction. The findings suggest that cultivating a varied and vibrant cultural existence, which encompasses activities such as traveling, engaging in cultural events, and being well-informed through reading, can greatly augment a person's overall life contentment. Policymakers, communities, and individuals can promote these cultural activities as part of efforts to improve well-being and quality of life.

*5.3.3. Implications of physical factors*

The physical factors identified in our LifeWell survey, including age, self-rated health (A2), long-term physical health issues or disabilities (C1), height (E1), weight (E2), and consultations with practitioners or therapists (E5_a), are crucial for evaluating life satisfaction. These features collectively capture the essence of physical well-being, significantly impacting overall life satisfaction. For instance, age affects life satisfaction through its implications on health, cognitive function, and social roles. Similarly, poor self-rated health and chronic physical conditions limit daily activities and independence. Furthermore, examining height and weight provides insights into how physical characteristics, influenced by societal norms and personal body image, affect life satisfaction. This holistic view underscores the intricate link between physical health and life satisfaction, stressing the need for comprehensive well-being assessments.

*5.3.4. Implications of mental health-related factors*

The extracted features related to depression (D2), stress management (D4, D6, D8, D10), perseverance (D11, D15), and neuroticism (D16, D17) have important implications for an individual's mental health and overall life satisfaction. The identification of these factors can help individuals become more self-aware of their psychological strengths and weaknesses. This self-knowledge can empower them to seek appropriate support and interventions to improve their mental well-being and life satisfaction. Therapists and psychiatrists can use this information to develop more targeted and personalized treatment plans for their clients. By understanding the specific psychological factors impacting life satisfaction, they can design tailored therapy approaches to address issues like depression, stress management, and personality traits. Healthcare providers can leverage this finding to screen patients more comprehensively for mental health concerns. At a broader societal level, these findings can inform public health initiatives and policies aimed at improving mental health and life satisfaction. Governments can use this information to allocate resources for mental health education, community-based support programs, and accessible therapies targeted at addressing the specific psychological factors identified.

*5.4. Limitations*

Despite the promising findings of our research and the robust performance of the machine learning models in predicting life satisfaction, the following concerns remain:

- Similar to previous works in life satisfaction prediction, the generalizability of the predictive models is constrained due to the singular cultural and socioeconomic context of the dataset, which originates from Denmark and encompasses individuals aged 16-64. As a result, the models' applicability may not extend seamlessly across different populations and cultures. The dataset's demographic specificity may also limit the model's applicability and generalizability across different cultural, socioeconomic, and age demographics. The potential for broader application of the methods is limited by the scope of the dataset.
- The model might not account for the temporal and dynamic nature of life satisfaction by relying on static, cross-sectional data. Inherent fluctuations and evolving circumstances that could impact life satisfaction over time are not encapsulated within the present model.

## 6. Conclusion and future work

The study of life satisfaction is essential for ensuring human well-being. Existing methods for measuring life satisfaction come with validation and propagation concerns. Most measurement schemes of life satisfaction involve complex statistical analysis, large-scale surveying, and erroneous and unexplained predictions.

The machine learning model used in this study has successfully predicted the psychological state of our subjects with an accuracy of 93.80% and macro F1-score of 73.00%, which requires a survey of only 27 questions. Such a small set of questions makes the survey process simple and easily reproducible. It also gives psychologists insight into the right questions to ask their patients to assess their mental state. Additionally, suppose there is any concern about the justification of the decisions made by the machine learning model. In that case, the explanations provided by the Explainable AI should suffice to offer credibility to the decision-making process. We, therefore, conclude that the employment of machine learning and explainable AI can significantly reduce errors and complexity in predicting subjects' state of life satisfaction. We further show the logical correlation of the extracted indicators with different populations based on several age brackets and provide insights based on physical, social, economic, and cultural grounds.

In addition to the machine learning models, large language models (LLMs) were explored for life satisfaction prediction. By converting the tabular data into natural language sentences through mapping and adding meaningful counterparts, an accuracy of 93.74% and macro F1-score of 73.21% was achieved using the BioBERT model. Furthermore, ablation studies were conducted to understand the impact of data resampling and feature selection techniques on the overall model performance.

For future works, we plan to validate and possibly enhance the models' robustness by incorporating data from a diverse array of countries, thus ensuring a more universally applicable predictive framework. We plan on expanding the study to multifarious demographic and cultural contexts and comprehensively appraising the generalizability of the findings across heterogeneous populations. We further plan to evaluate the approach by implementing deeper neural network architectures. Deep learning models, especially those employing recurrent neural networks (RNNs) or transformers, could offer advanced capability in capturing temporal dependencies and complex patterns within longitudinal life satisfaction data, potentially uncovering deeper insights and novel relationships among variables.

**Ethics statement**

Review and/or approval by an ethics committee was not needed for this study because the dataset used in this study is publicly available to researchers.

Informed consent was not required for this study because the data were anonymized and de-identified prior to analysis, ensuring that the privacy and confidentiality of the individuals whose data were included in the dataset are maintained. Additionally, the study design did not involve any interaction with human subjects or have any impact on their rights and welfare.

**CRediT authorship contribution statement**

**Alif Elham Khan:** Writing – review & editing, Writing – original draft, Visualization, Software, Methodology, Investigation, Formal analysis, Conceptualization. **Mohammad Junayed Hasan:** Writing – review & editing, Writing – original draft, Visualization, Software, Methodology, Investigation, Formal analysis, Conceptualization. **Humayra Anjum:** Writing – review & editing, Writing – original draft, Visualization, Software, Methodology, Investigation, Conceptualization. **Nabeel Mohammed:** Writing – original draft,

Supervision, Project administration, Investigation. **Sifat Momen:** Writing – review & editing, Writing – original draft, Supervision, Project administration, Investigation, Formal analysis, Conceptualization.

**Declaration of competing interest**

The authors declare that they have no known competing financial interests or personal relationships that could have appeared to influence the work reported in this paper.

**Data availability statement**

The dataset utilized in this article is a publicly available dataset. It is directly accessible via the Dryad repository at https://doi.org/10.5061/dryad.qd2nj.